%% file: _main.tex
\input{_constants}
\arxiv

\pdfoutput=1
\documentclass[10pt,twocolumn,letterpaper]{article}
\input{cvpr_header}

\begin{document}
\title{\paperTitle}
\author{\authorBlock}
\maketitle

\input{00_abstract}
\input{01_intro}
\input{02_related}

\input{03_preliminary}
\input{04_method}

\input{05_experiments}

\input{10_conclusion}

{\small
\bibliographystyle{ieee_fullname}
\bibliography{11_references}
}

\ifarxiv \clearpage \input{12_appendix} \fi

\end{document}

%% file: _constants.tex
\def\cvprPaperID{2584}
\def\confName{CVPR}
\def\confYear{2023}

\def\paperTitle{Learning Attention as Disentangler for Compositional Zero-shot Learning}

\def\authorBlock{
    Shaozhe Hao \qquad
    Kai Han\thanks{Corresponding author} \qquad
    Kwan-Yee K. Wong \\
    The University of Hong Kong\\
    {\tt\small \{szhao, kykwong\}@cs.hku.hk \qquad kaihanx@hku.hk}
}

\newif\ifreview 
\newif\ifarxiv \newcommand{\arxiv}{\arxivtrue}
\newif\ifcamera 
\newif\ifrebuttal 

%% file: cvpr_header.tex
\ifreview \usepackage[review]{cvpr} \fi
\ifarxiv \usepackage[pagenumbers]{cvpr} \fi
\ifrebuttal \usepackage[rebuttal]{cvpr} \fi
\ifcamera \usepackage{cvpr} \fi

\usepackage{graphicx}
\usepackage{amsmath}
\usepackage{amssymb}
\usepackage{booktabs}

\input{_macros}  

\usepackage{xr-hyper}

\makeatletter
\newcommand*{\addFileDependency}[1]{
  \typeout{(#1)}
  \@addtofilelist{#1}
  \IfFileExists{#1}{}{\typeout{No file #1.}}
}

\makeatother

\usepackage[pagebackref,breaklinks,colorlinks]{hyperref}
\usepackage[capitalize]{cleveref}
\crefname{section}{Sec.}{Secs.}
\crefname{table}{Table}{Tables}
\crefname{figure}{Fig.}{Figs.}

\hypersetup{
    colorlinks=true,
    linkcolor=BrickRed,
    filecolor=cyan,      
    urlcolor=magenta,
    citecolor=Green,
} 

%% file: _macros.tex
\usepackage{times}
\usepackage{microtype}
\usepackage{epsfig}
\usepackage[dvipsnames,table,xcdraw]{xcolor}
\usepackage{caption}
\usepackage{float}
\usepackage{placeins}
\usepackage{color, colortbl}
\usepackage{stfloats}
\usepackage{enumitem}
\usepackage{tabularx}
\usepackage{xstring}
\usepackage{multirow}
\usepackage{xspace}
\usepackage{url}
\usepackage{soul}
\usepackage{subcaption}
\usepackage{xcolor}
\usepackage[hang,flushmargin]{footmisc}
\usepackage{amssymb}
\usepackage{pifont}
\usepackage[accsupp]{axessibility}

\newcommand{\cmark}{\ding{51}}%
\newcommand{\xmark}{\ding{55}}%

\makeatletter
\renewcommand{\paragraph}{%
  \@startsection{paragraph}{4}%
  {\z@}{1ex \@plus .2ex \@minus .2ex}{-1em}%
  {\normalfont\normalsize\bfseries}%
}
\makeatother

\ifcamera \usepackage[accsupp]{axessibility} \fi

% ...

\ifarxiv  \fi

\newcommand{\R}[1]{{%
    \textbf{%
        \ifstrequal{#1}{hmSm}{\textcolor{red}{#1}}{%
        \ifstrequal{#1}{1}{\textcolor{red}{R#1}}{%
        \ifstrequal{#1}{GrqZ}{\textcolor{blue}{#1}}{%
        \ifstrequal{#1}{2}{\textcolor{blue}{R#1}}{%
        \ifstrequal{#1}{GTBY}{\textcolor{teal}{#1}}{%
        \ifstrequal{#1}{3}{\textcolor{teal}{R#1}}{%
        \ifstrequal{#1}{4}{\textcolor{magenta}{R#1}}{%
                           \textcolor{cyan}{R#1}%
        }}}}}}}%
    }%
}}

\def\eg{\emph{e.g.}}

\def\ie{\emph{i.e.}}
\def\etal{\emph{et al.}}

\definecolor{attr}{RGB}{192,0,0}
\definecolor{comp}{RGB}{0,164,74}
\definecolor{obj}{RGB}{0,112,192}
\definecolor{tabcolor}{RGB}{237,244,254}
\definecolor{branchone}{RGB}{85,131,53}
\definecolor{branchtwo}{RGB}{197,91,17}
\newcommand{\framework}{ADE\xspace}
\newcommand{\hsznew}[1]{#1}
\newcommand{\hsz}[1]{#1}

\usepackage[ruled,vlined]{algorithm2e}
\definecolor{commentcolor}{RGB}{110,154,155}   
\newcommand{\PyComment}[1]{\ttfamily\textcolor{commentcolor}{\# #1}}  
\newcommand{\PyCode}[1]{\ttfamily\textcolor{black}{#1}} 

\usepackage[normalem]{ulem}
\newcommand{\li}{\uline{\hspace{0.5em}}}

%% file: 00_abstract.tex
\begin{abstract}
Compositional zero-shot learning (CZSL) aims at learning visual concepts (\ie, attributes and objects) from seen compositions and combining concept knowledge into unseen compositions. The key to CZSL is learning the disentanglement of the attribute-object composition. To this end, we propose to exploit cross-attentions as compositional disentanglers to learn disentangled concept embeddings. For example, if we want to recognize an unseen composition ``yellow flower", we can learn the attribute concept ``yellow" and object concept ``flower" from different yellow objects and different flowers respectively. To further constrain the disentanglers to learn the concept of interest, we employ a regularization at the attention level. Specifically, we adapt the earth mover's distance (EMD) as a feature similarity metric in the cross-attention module. Moreover, benefiting from concept disentanglement, we improve the inference process and tune the prediction score by combining multiple concept probabilities. Comprehensive experiments on three CZSL benchmark datasets demonstrate that our method significantly outperforms previous works in both closed- and open-world settings, establishing a new state-of-the-art. Project page: 
\url{https://haoosz.github.io/ade-czsl/}
\end{abstract}

%% file: 01_intro.tex
\section{Introduction}
\label{sec:intro}
Suppose we have never seen white bears (\ie, polar bears) before. Can we picture what it would look like? This is not difficult because we have seen many white animals in daily life (\eg, white dogs and white rabbits) and different bears with various visual attributes in the zoo (\eg, brown bears and black bears). Humans have no difficulty in disentangling ``white" and ``bear" from seen instances and combining them into the unseen composition. Inspired by this property of human intelligence, researchers attempt to make machines learn compositions of concepts as well. Compositional zero-shot learning (CZSL) is a specific problem studying visual compositionality, aiming to learn visual concepts from seen compositions of attributes and objects and generalize concept knowledge to unseen compositions.

\begin{figure}[t]
    \centering
    \includegraphics[width=\linewidth]{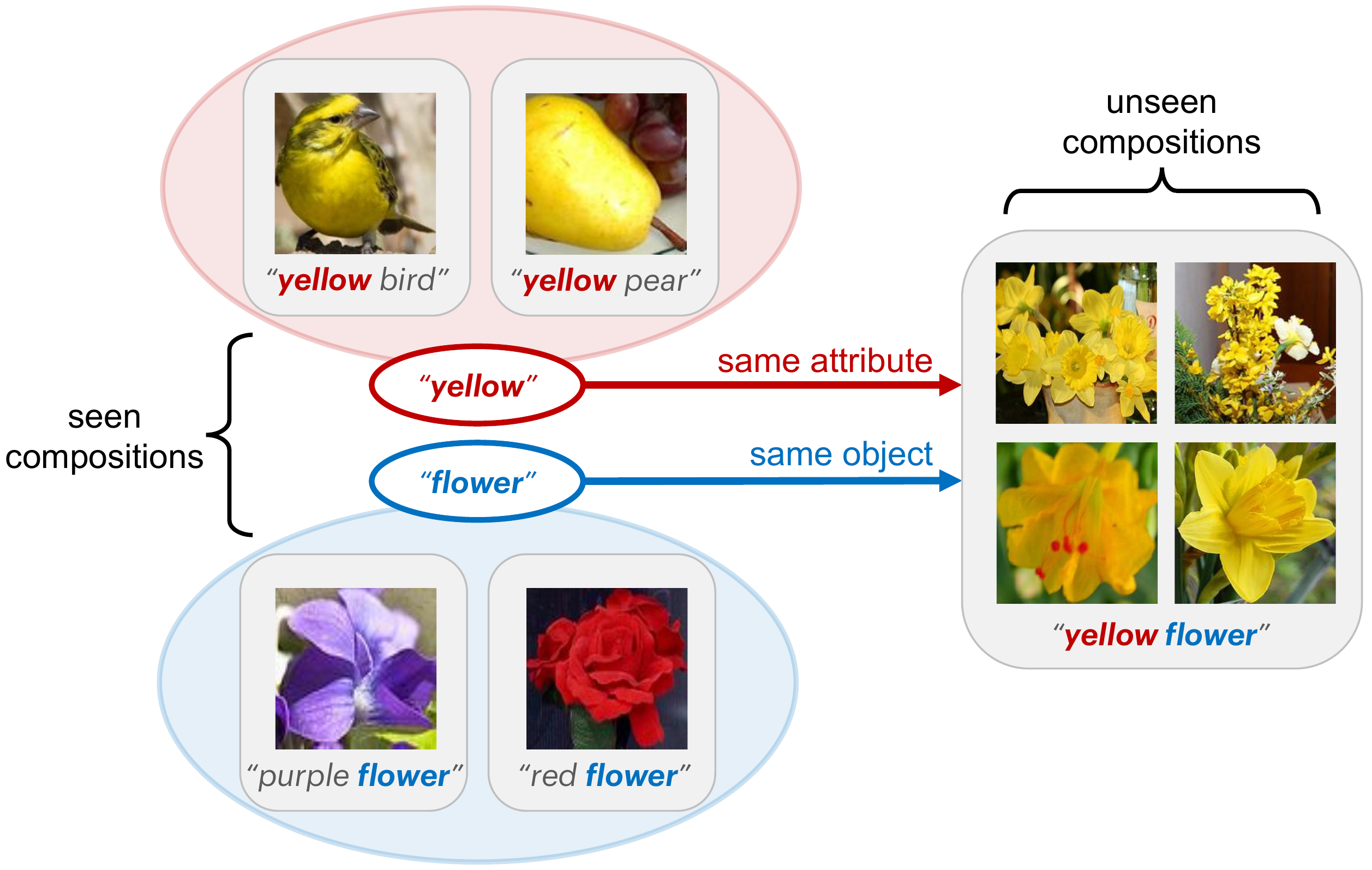}
    \caption{Motivation illustration. Given images from seen attribute-object compositions, human can disentangle the attribute ``yellow" from ``yellow bird" and ``yellow pear", and the object ``flower" from ``purple flower" and ``red flower". After learning visual properties of the concepts ``yellow" and ``flower", human can then recognize images from the unseen composition ``yellow flower".}
    \label{fig:overview}
    \vspace{-3pt}
\end{figure}

Learning attribute-object compositions demands prior knowledge about attributes and objects. However, visual concepts of attributes and objects never appear alone in a natural image. To learn exclusive concepts for compositionality learning, we need to disentangle the attribute concept and the object concept. As illustrated in~\cref{fig:overview}, if we want to recognize the image of ``yellow flower", it is necessary to learn the ``yellow" concept and the ``flower" concept, \ie, disentangle visual concepts, from images of seen compositions. Previous works~\cite{misra2017red,nagarajan2018attributes,purushwalkam2019task,wei2019adversarial,li2020symmetry,naeem2021learning,mancini2021open,mancini2022learning} tackle CZSL by composing attribute and object word embeddings, and projecting word and visual embeddings to a joint space. They fail to disentangle visual concepts.
Recently, some works~\cite{ruis2021independent,zhang2022learning,li2022siamese, Saini_2022_CVPR} consider visual disentanglement but still have limitations despite their good performance. SCEN~\cite{li2022siamese} learns concept-constant samples contrastively without constructing concept embedding prototypes to avoid learning irrelevant concepts shared by positive samples. IVR~\cite{zhang2022learning} disentangles visual features into ideal concept-invariant domains. This ideal domain generalization setting requires a small discrepancy of attribute and object sets and would degenerate on vaster and more complex concepts. ProtoProp~\cite{ruis2021independent} and OADis~\cite{Saini_2022_CVPR} learn local attribute and object prototypes from spatial features on convolutional feature maps. However, spatial disentanglement is sometimes infeasible because attribute and object concepts are highly entangled in spatial features. Taking an image of ``yellow flower" as an example, the spatial positions related to the attribute ``yellow" and the object ``flower" completely overlap, which hinders effective attribute-object disentanglement.

To overcome the above limitations, we propose a simple visual disentangling framework exploiting \textbf{A}ttention as \textbf{D}is\textbf{E}ntangler (\framework) on top of vision transformers~\cite{dosovitskiy2020vit}. We notice that vision transformers (ViT) have access to more sub-space global information across multi-head attentions than CNNs~\cite{raghu2021vision}. Therefore, with the expressivity of different subspace representations, token attentions of ViT may provide a more effective way for disentangling visual features, compared to using traditional spatial attentions across local positions on convolutional features~\cite{Saini_2022_CVPR, ruis2021independent}. Specifically, it is difficult to disentangle the attribute-object composition ``yellow flower" by spatial positions, but it is possible for ViT multi-head attentions to project attribute concept ``yellow" and object concept ``flower" onto different subspaces. Inspired by this property, we propose to learn cross-attention between two inputs that share the same concept, \eg, ``yellow bird" and ``yellow pear" share the same attribute concept ``yellow". In this way, we can derive attribute- and object-exclusive visual representations by cross-attention disentanglement. 
To ensure that the concept disentangler is exclusive to the specific concept, we also need to constrain the disentanglers to learn the concept of interest instead of the other concept. For example, given attribute-sharing images, the attribute attention should output similar attribute-exclusive features while the object attention should not. To achieve this goal, we apply a regularization term adapted from the earth mover's distance (EMD)~\cite{emd_paper} at the attention level. This regularization term forces cross-attention to learn the concept of interest by leveraging the feature similarity captured from all tokens. 
Mancini \etal~\cite{mancini2021open} propose an open-world evaluation setting, which is neglected by most previous works. We consider both closed-world and the open-world settings in our experiments, demonstrating that our method is coherently efficient in both settings. 
The contributions of this paper are summarized below: 
\begin{itemize}[itemsep=2pt,topsep=0pt,parsep=0pt,leftmargin=15pt]
    \item We propose a new CZSL approach, named \framework, using cross-attentions to disentangle attribute- and object-exclusive features from paired concept-sharing inputs.
    \item We force attention disentanglers to learn the concept of interest with a regularization term adapted from EMD, ensuring valid attribute-object disentanglement.
    \item We comprehensively evaluate our method in both closed-world and open-world settings on three CZSL datasets, achieving consistent state-of-the-art.
\end{itemize}

%% file: 02_related.tex
\section{Related work}
\label{sec:related}
\emph{Visual attribute} has been widely studied to understand how visual properties can be learned from objects.  The pioneering work by Ferrari and Zisserman~\cite{ferrari2007learning} learned visual attributes using a probabilistic generative model. The successive work by Lampert \etal~\cite{lampert2009learning} used visual attributes to detect unseen objects with an attribute-based multi-label classification. Similarly, Patterson \etal~\cite{patterson2016coco} proposed Economic Labeling Algorithm (ELA) to discover multi-label attributes for objects. Different from multi-label classification, other works~\cite{chen2014inferring,hwang2011sharing,farhadi2009describing,mahajan2011joint} learned attribute-object relationship to generalize attribute feature across all object categories based on probabilistic models. Visual attributes also benefit downstream tasks, \eg, object recognition~\cite{nan2019recognizing,isola2015discovering,farhadi2010attribute}, action recognition~\cite{alayrac2017joint,fathi2013modeling,mccandless2013object}, image captioning~\cite{ordonez2011im2text,kulkarni2013babytalk}, and semi-supervised learning~\cite{shrivastava2012constrained}.

\emph{Compositional zero-shot learning (CZSL)} is a special case of zero-shot learning (ZSL)~\cite{palatucci2009zero,xian2017zero,xian2018feature,socher2013zero,romera2015embarrassingly}, aims at recognizing unseen attribute-object compositions learning from seen compositions. Misra \etal~\cite{misra2017red} first termed and studied CZSL by projecting composed primitives and visual features to a joint embedding space. Nagarajan \etal~\cite{nagarajan2018attributes} formulated attributes as matrix operators applied on object vectors. Purushwalkam \etal~\cite{purushwalkam2019task} introduced a task-driven modular architecture to learn unseen compositions by re-weighting a set of sub-tasks. Wei \etal~\cite{wei2019adversarial} generated attribute-object compositions with GAN~\cite{goodfellow2014generative} to match visual features. Li \etal~\cite{li2020symmetry} proposed symmetry principle of attribute-object transformation under the supervision of group axioms. Naeem \etal~\cite{naeem2021learning} and Mancini \etal~\cite{mancini2022learning} used graph convolutional networks to extract attribute-object representations. Recently, some works shift their interest from word composing to visual disentanglement. Atzmon \etal~\cite{atzmon2020causal} solved CZSL from a causal perspective to learn disentangled representations. Ruis \etal~\cite{ruis2021independent} proposed to learn prototypical representations of objects and attributes. Li \etal~\cite{li2022siamese} disentangled visual features into a Siamese contrastive space and entangled them with a generative model. Saini \etal~\cite{Saini_2022_CVPR} extracted visual similarity from spatial features to disentangle attributes and objects. Zhang \etal~\cite{zhang2022learning} treated CZSL as a domain generalization task, learning attribute- and object-invariant domains. A more realistic open-world CZSL setting was studied in \cite{mancini2021open,mancini2022learning,karthik2022kg}, which considered all possible compositions in testing. Very recently, an inspiring work by Nayak \etal~\cite{nayak2022learning} introduced compositional soft prompts to CLIP~\cite{radford2021learning} to tackle CZSL problem.

\emph{Attention mechanism} has been well studied by non-local neural networks~\cite{wang2018non} in computer vision and transformers~\cite{vaswani2017attention} in machine translation. Dosovitskiy \etal~\cite{dosovitskiy2020vit} adapted transformer architecture to computer vision field, proving its comparable efficiency over traditional CNNs. Inspired by the multi-head self-attention implemented by transformers, our work exploits efficient attention as disentangler for CZSL.

%% file: 03_preliminary.tex
\begin{figure*}[t]
    \centering
    \includegraphics[width=\linewidth]{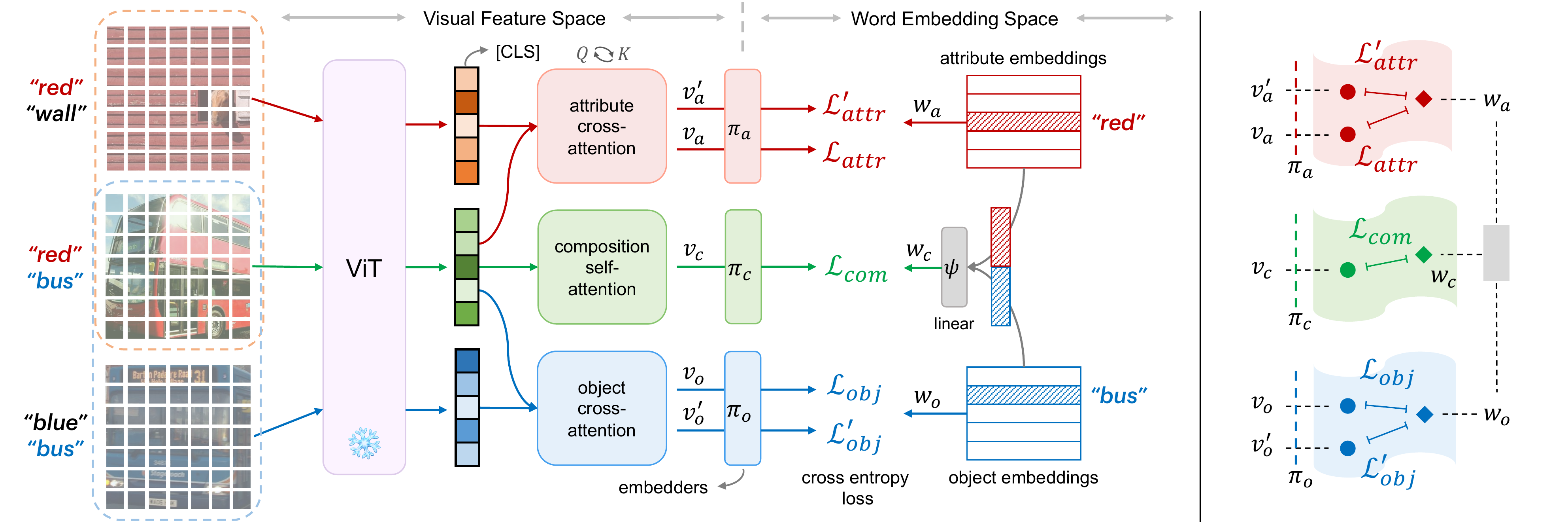}
    \caption{Method overview. Left (our framework \framework): Given one target image of ``red bus", we sample two auxiliary images of the same attribute ``red wall" and of the same object ``blue bus". We feed the three images into a frozen ViT initialized with DINO~\cite{caron2021emerging}. We then input all encoded tokens (\ie, \texttt{[CLS]} and patch tokens) to three attention modules: (1) attribute cross-attention taking paired attribute-sharing tokens as inputs; (2) object cross-attention taking paired object-sharing tokens as inputs; (3) composition self-attention taking tokens of single target image as input. We then project the \texttt{[CLS]} tokens of attention outputs with three MLP embedders $\pi_a$, $\pi_c$, and $\pi_o$. We finally compute cross entropy losses of embedded visual features with three learnable word embeddings: attribute embeddings, object embeddings, and their linear fused composition embeddings. Right: Illustration of five losses in attribute, object, and composition embedding spaces.}
    \label{fig:pipeline}
\end{figure*}

\section{Preliminary}
\label{sec:preliminary}
\vspace{-3pt}
Compositional zero-shot learning aims at learning a model from limited compositions of attributes (\eg, yellow) and objects (\eg, flower) to recognize an image from novel compositions. Given a set of all possible attributes $\mathcal{A}=\{a_0, a_1, \cdots, a_n\}$ and a set of all possible objects $\mathcal{O}=\{o_0, o_1, \cdots, o_m\}$, a compositional class set $\mathcal{C} = \mathcal{A} \times \mathcal{O}$ includes all attribute-object compositions. We divide $\mathcal{C}$ into two disjoint sets, namely the set of seen classes $\mathcal{C}_{s}$ and the set of unseen classes $\mathcal{C}_{u}$, where $\mathcal{C}_{s} \cap \mathcal{C}_{u} = \varnothing$ and $\mathcal{C}_{s} \cup \mathcal{C}_{u} = \mathcal{C}$. Training images are only from classes in $\mathcal{C}_{s}$ and testing images are from classes in $\mathcal{C}_{test} = \mathcal{C}_{s} \cup \mathcal{C}_{u}^\prime$. For closed-world evaluation, $\mathcal{C}_{u}^\prime$ is a predefined subset of $\mathcal{C}_{u}$, \ie, $\mathcal{C}_{u}^\prime \subset \mathcal{C}_{u}$. For open-world evaluation, all possible compositions are considered for testing, \ie, $\mathcal{C}_{u}^\prime = \mathcal{C}_{u}$. CZSL task aims to learn a model $f: \mathcal{X} \rightarrow \mathcal{C}_{test}$ to predict labels in the testing composition set $c \in \mathcal{C}_{test}$ for input images $x \in \mathcal{X}$.

To learn a CZSL model, a common method is to minimize the cosine similarity score between the visually encoded feature and the word embedding of attribute-object compositions. The similarity score can be expressed as
\begin{equation}
\setlength{\abovedisplayskip}{3pt}
\setlength{\belowdisplayskip}{3pt}
    s(x, (a, o)) = \frac{\phi(x)}{||\phi(x)||} \cdot \frac{\psi(a,o)}{||\psi(a,o)||}
\end{equation}
where $\phi(\cdot)$ is the visual encoder, and $\psi(\cdot, \cdot)$ is the composition function. The CZSL model $f$ can be formulated as
\begin{equation}
\setlength{\abovedisplayskip}{3pt}
\setlength{\belowdisplayskip}{3pt}
    f(x; \theta_\phi, \theta_\psi) = \mathop{\arg\max}_{a_i \in \mathcal{A}, o_j \in \mathcal{O}} s(x, (a_i, o_j))
\end{equation}
The model $f$ is optimized by learning the model parameters $\theta_\phi$ and $\theta_\psi$ to match the input $x$ with the most similar attribute-object composition. Visual encoder $\phi$ and composition function $\psi$, as two core components, take multiple forms in different methods. For visual encoder, works~\cite{ruis2021independent, zhang2022learning, Saini_2022_CVPR} manually design networks on top of a pretrained backbone (\eg, ResNet18~\cite{he2016deep}) to disentangle attribute and object embeddings for visual features. For composition functions, the object conditioned network~\cite{Saini_2022_CVPR}, the graph embedding~\cite{naeem2021learning, ruis2021independent}, and the linear projector~\cite{mancini2021open, zhang2022learning} are commonly used. Unlike employing complex composition functions, we adopt the simple linear projector in our method. 

%% file: 04_method.tex
\section{Our approach}
\label{sec:method}
We aim to disentangle attribute and object features for visual representation by learning from image pairs sharing the same attributes or objects. For example, by learning the object ``flower" from the ``purple flower" and the ``red flower" and learning the attribute ``yellow" from the ``yellow bird" and the ``yellow pear", we can infer what a ``yellow flower" looks like. To this end, we propose ADE to learn cross-attentions as concept disentanglers with the regularization of a novel token earth mover's distance. The whole framework of our method is shown in~\cref{fig:pipeline}.

\subsection{Disentanglement with cross-attention}
Multi-head self-attention in transformers~\cite{vaswani2017attention} is powerful for extracting token connections. Attention maps each query token, and key-value token pairs to an output token. Every output token is a weighted sum of the values, where the weights is the similarity of the corresponding query token and the key tokens. The attention is formulated as:
\begin{equation}
    Attention(Q, K, V) = softmax(\frac{QK^T}{\sqrt{d_k}}) V
    \label{eq:attention}
\end{equation}
where three inputs are query tokens ($Q$), key tokens ($K$), and value tokens ($V$) and the scaling factor $d_k$ is the dimension of the query and key. Self-attention uses the same input for $Q$, $K$, $V$ and effectively captures relationships among \verb+[CLS]+ and patch tokens of a single input image. Normally, the \texttt{[CLS]} token is exclusively used as a global representation of the input image for downstream tasks, but exclusive \texttt{[CLS]} token highly entangles attribute and object concepts.

\begin{figure}[t]
    \centering
    \includegraphics[width=\linewidth]{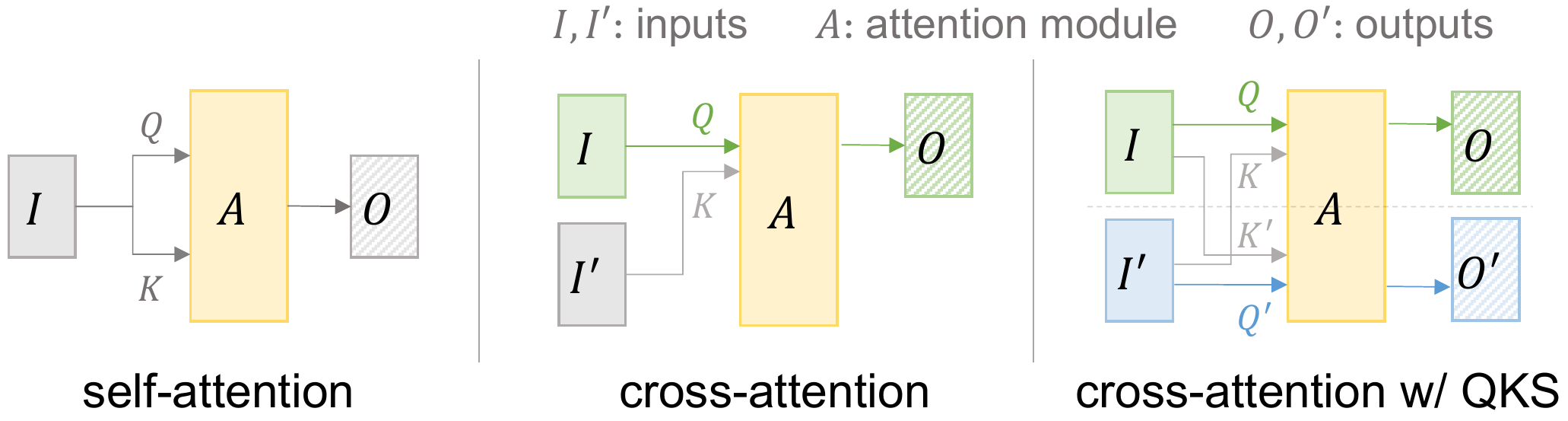}
    \caption{Illustration of three types of attention.}
    \label{fig:compare}
\end{figure}

\hsz{For CZSL task, we want to disentangle the exclusive concept by exploiting the semantic similarity between tokens from different concept-sharing features.} For this purpose, we introduce a \emph{cross}-attention mechanism to extract attribute-exclusive or object-exclusive features for paired images sharing the same attribute or object concept. The cross-attention has the same structure as the self-attention, but works on different inputs. Self-attention uses the same input $I$ for $Q$, $K$, $V$ (\ie, $Q$=$K$=$V$=$I$) while cross-attention uses one of the paired inputs $I$ for the query and the other $I^{\prime}$ for the key and the value (\ie, $Q$=$I$, $K$=$V$=$I^{\prime}$). 
In the cross-attention, the query input and the key input play different roles. 
The cross-attention output is the weighted sum of the value (also the key) input based on the similarity between the query and key inputs. For a pair of input images, we swap the query and key inputs in computing cross-attention to derive two cross-attended outputs for the respective inputs. Cross-attention with query-key swapping (QKS) is illustrated in~\cref{fig:compare}. 

The cross-attention outputs paired features of the exclusive concept with paired concept-sharing inputs. Taking the object-sharing pair as an example, object cross-attention leverages the shared object features of the key input to enhance the query input, thus making the outputs exclusive to object feature space, \ie, object-exclusive features $v_o$ and $v_o^{\prime}$. The same applies to attribute-exclusive features $v_a$ and $v_a^{\prime}$ extracted from paired attribute-sharing images. We also use a standard self-attention to extract composition feature $v_c$.

To train cross-attentions, we adopt common cross entropy loss after MLP embedders ($\pi_a$, $\pi_c$, $\pi_o$) for attribute-exclusive features ($v_a$, $v_a^\prime$), object-exclusive features ($v_o$, $v_o^\prime$), and composition feature ($v_c$). \hsz{We denote any of these visual features as $v$ and the corresponding embedder as $\pi$. We denote the word embedding as $w_z$, where the concept $z \in \mathcal{Z}$ can be any of the attribute $a \in \mathcal{A}$, the object $o \in \mathcal{O}$, or the composition $c \in \mathcal{C}$. The class probability of L2-normalized $\pi(v)$ and $w_z$ can be computed as:}
\begin{equation}
\setlength{\abovedisplayskip}{3pt}
\setlength{\belowdisplayskip}{3pt}
     p_\pi(z \ |\ v)  = \frac{\exp{(\pi(v) \cdot w_z/\tau)}}{\sum_{\hat{z} \in \mathcal{Z}} \exp{(\pi(v) \cdot w_{\hat{z}}/\tau)}}, z \in \mathcal{Z}
\end{equation}
where $\tau$ is the temperature.  \hsz{We then compute the cross entropy to classify visual features:}
\begin{equation}
\setlength{\abovedisplayskip}{3pt}
\setlength{\belowdisplayskip}{3pt}
    H_\pi(v, z) = - \log p_\pi(z \ |\ v), z \in \mathcal{Z}
    \label{eq:cross-entropy}
\end{equation}
\hsz{Considering different classification objectives of attribute, object, and composition, the general~\cref{eq:cross-entropy} can be applied to five visual features, \ie, $v_a$, $v_a^\prime$, $v_o$, $v_o^\prime$, $v_c$.} Therefore, the training objective is composed of five cross entropy losses:
{\setlength{\abovedisplayskip}{0pt}
\begin{align}
    \mathcal{L}_{ce} = \ & \textcolor{attr}{\underbrace{\textcolor{black}{H_{\pi_a}(v_a, a)}}_{\mathcal{L}_{attr}}} + \textcolor{attr}{\underbrace{\textcolor{black}{H_{\pi_a}(v_a^\prime, a)}}_{\mathcal{L}_{attr}^\prime}} +  \textcolor{comp}{\underbrace{\textcolor{black}{H_{\pi_c}(v_c, c)}}_{\mathcal{L}_{com}}} + \notag \\ 
    &  \textcolor{obj}{\underbrace{\textcolor{black}{H_{\pi_o}(v_o, o)}}_{\mathcal{L}_{obj}}} + \textcolor{obj}{\underbrace{\textcolor{black}{H_{\pi_o}(v_o^\prime, o)}}_{\mathcal{L}_{obj}^\prime}}
\end{align}
}where the attribute label $a \in \mathcal{A}$, the object label $o \in \mathcal{O}$, and the composition label $c=(a, o) \in \mathcal{C}_{test}$.

\subsection{Disentangling constraint at the attention level}
\hsz{So far, we have enabled cross-attentions to disentangle concepts, but we want to further ensure that the attribute and the object disentanglers learn the corresponding attribute and object concepts instead of the opposite concepts. To this end, we introduce an attention-level earth mover's distance (EMD) to constrain disentanglers to learn the concept of interest.} The EMD is formulated as an optimal transportation problem in~\cite{emd_paper}. Suppose we have supplies of $n_s$ sources $\mathcal{S}=\{s_i\}_{i=1}^{n_s}$ and demands of $n_d$ destinations $\mathcal{D}=\{d_j\}_{j=1}^{n_d}$. Given the moving cost from the $i$-th source to the $j$-th destination $c_{ij}$, an optimal transportation problem aims to find the minimal-cost flow $f_{ij}$ from sources to destinations:
\begin{align}
    \underset{f_{ij}}{\text{minimize}} \quad &\sum\nolimits_{i=1}^{n_s}\sum\nolimits_{j=1}^{n_d} c_{ij} f_{ij} \\
    \text{subject to} \quad &f_{ij} \geqslant 0,\ i=1, ..., n_s,\ j= 1, ..., n_d \\
    &\sum\nolimits_{j=1}^{n_d} f_{ij} = s_i, \ i=1, ..., n_s \\
    &\sum\nolimits_{i=1}^{n_s} f_{ij} = d_j, \ j=1, ..., n_d
\end{align} 
where the optimal flow $\tilde{f}_{ij}$ is computed by the moving cost $c_{ij}$, the supplies $s_i$, and the demands $d_j$. The EMD can be further formulated as:
\begin{equation}
    \text{EMD}(c_{ij}, s_i, d_j)= (1-c_{ij})\tilde{f}_{ij}.
\end{equation}

\begin{figure}[t]
    \centering
    \includegraphics[width=\linewidth]{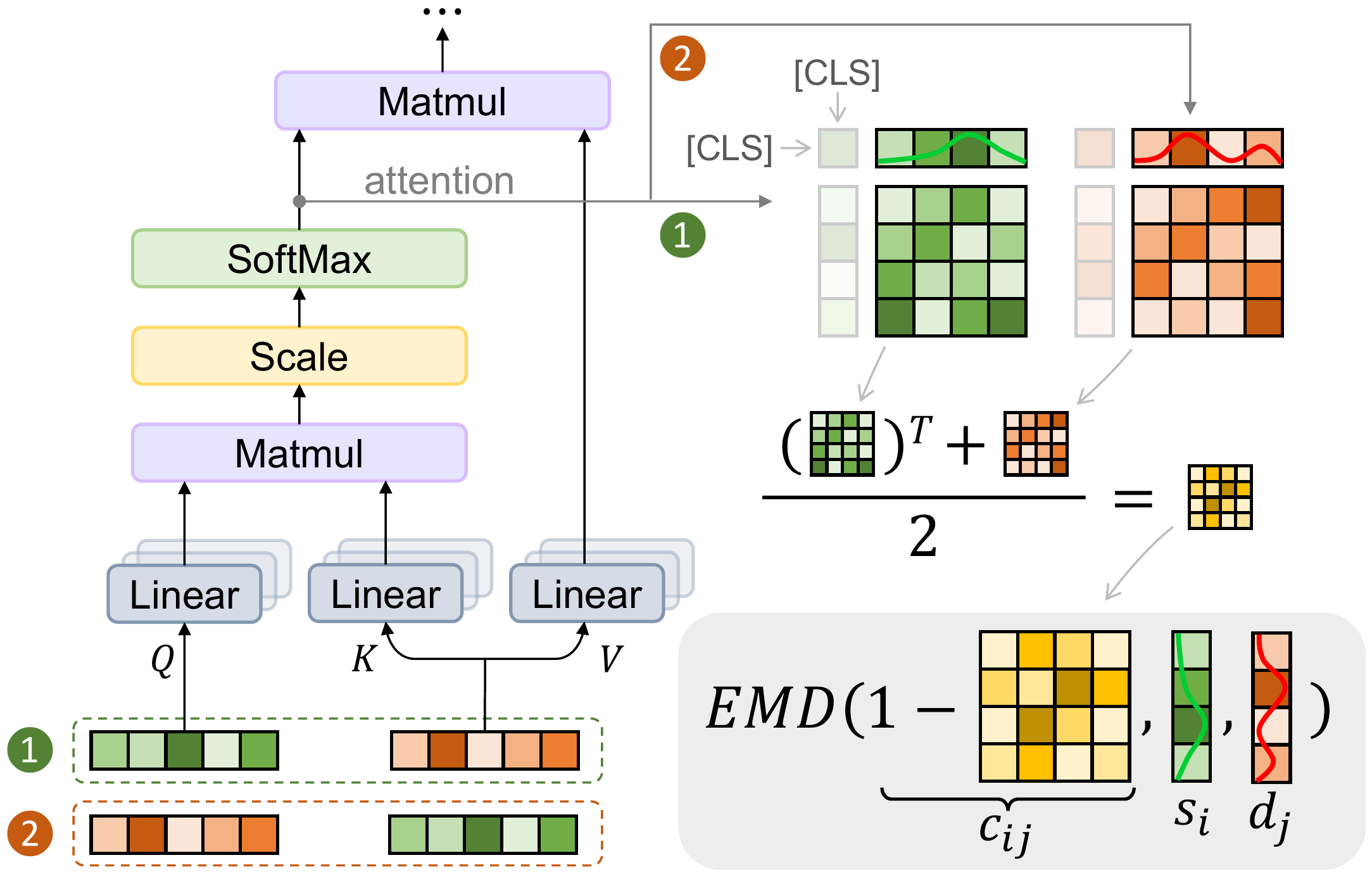}
    \caption{Illustration of the adapted EMD at the attention level. Branches \textcolor{branchone}{\ding{182}} and \textcolor{branchtwo}{\ding{183}} respectively denote using one of the paired inputs as the query $Q$. We compute the EMD with two attention maps from \textcolor{branchone}{\ding{182}} and \textcolor{branchtwo}{\ding{183}} branches. We use the \texttt{[CLS]}-to-patch logits as the supplies $s_i $ and the demands $d_j$, and one minus the mean matrix of patch-to-patch logits as the moving cost $c_{ij}$. In this way, we adapt the EMD to our cross-attention module.}
    \label{fig:emd}
\end{figure}

When the EMD is greater, the distributions $\mathcal{S}$ and $\mathcal{D}$ are closer. \hsz{In \cref{fig:emd}, we show how to use the EMD as a feature similarity metric at the attention level in our cross-attention module.} We can derive a cross-attention map from~\cref{eq:attention} and the counterpart cross-attention map after query-key swapping. We use the logits of \verb+[CLS]+-to-patch attentions as supplies $s_i$ and demands $d_j$, which represent how similar the global feature of one image is to the local patches of the other image. We use one minus the average map of patch-to-patch attentions as the moving cost $c_{ij}$. In this way, we can compute an adapted EMD $\lambda$ for the cross-attention:
\begin{equation}
\setlength{\abovedisplayskip}{5pt}
\setlength{\belowdisplayskip}{5pt}
    \lambda^p_q = \text{EMD}(c_{ij}, s_i, d_j; p, q), p, q \in \{a, o\}
\end{equation}
where $p$ represents the type of the cross-attention, \ie, attribute cross-attention when $p=a$ and object cross-attention when $p=o$, and $q$ represents the type of the inputs, \ie, attribute-sharing inputs when $q=a$ and object-sharing inputs when $q=o$. 

For attribute cross-attention, attribute-sharing inputs should have large EMD $\lambda^a_a$ while object-sharing inputs should have small EMD $\lambda^a_o$. It is opposite for object cross-attention, \ie, small $\lambda^o_a$ of attribute-sharing inputs and large $\lambda^o_o$ of object-sharing inputs. Thus, we formulate the regularization term as:
\begin{equation}
\setlength{\abovedisplayskip}{3pt}
\setlength{\belowdisplayskip}{3pt}
    \mathcal{L}_{reg} = \lambda^a_o + \lambda^o_a - \lambda^a_a - \lambda^o_o
    \label{eq:reg}
\end{equation}

\subsection{Training and inference}
At the training phase, we formulate our final loss as a cross entropy with regularization for all inputs:
\begin{equation}
\setlength{\abovedisplayskip}{3pt}
\setlength{\belowdisplayskip}{3pt}
    \mathcal{L} = \mathcal{L}_{ce} + \mathcal{L}_{reg}
\end{equation}

At inference phase, we feed the same test input into three attention branches in a self-attention manner, deriving three class probabilities $p(c)$, $p(a)$, and $p(o)$\footnote{Abbreviations for $p_{\pi_c}(c\ |\ v_c)$, $p_{\pi_a}(a\ |\ v_a)$, and $p_{\pi_o}(o\ |\ v_o)$, where $v_c$, $v_a$, and $v_o$ are outputs from three self-attentions.}, where $c=(a, o)$. Unlike most methods modeling single $p(c)$ for prediction, we compute our prediction score by synthesizing attribute, object and composition probabilities:
\begin{equation}
    \hat{c} = \mathop{\arg\max}_{c \in \mathcal{C}_{test}} \ p(c) + \beta \cdot p(a) \cdot p(o)
    \label{eq:predict}
\end{equation}
We first fix $\beta$ = 1.0 during training and then validate $\beta$ = 0.0, 0.1, $\cdots$, 1.0 to choose the best $\beta$ on the validation set. Finally, we compute the composition prediction as~\cref{eq:predict} with the chosen $\beta$, making the best prediction trade-off between generalized composition and independent concept.

%% file: 05_experiments.tex
\section{Experiments}
\label{sec:experiments}
\subsection{Experimental details}
\paragraph{Datasets} We use three benchmark datasets in CZSL problem, namely Clothing16K~\cite{zhang2022learning}, UT-Zappos50K~\cite{yu2014fine}, and C-GQA~\cite{naeem2021learning}. Clothing16K~\cite{zhang2022learning} contains different types of clothing (\eg, shirt, pants) with color attributes (\eg, white, black). UT-Zappos50K~\cite{yu2014fine} is a fine-grained dataset consisting of different kinds of shoes (\eg, sneakers, sandals) with texture attributes (\eg, leather, canvas). C-GQA~\cite{naeem2021learning} is a split built on top of Stanford GQA dataset~\cite{hudson2019gqa}, composed of extensive common attribute concepts (\eg, old, wet) and object concepts (\eg, dog, bus) in real life. We follow the common data splits of these three datasets 
(see~\cref{tab:data-splits}). 

\begin{table}[h]
    \centering
    \scalebox{0.54}{
    \begin{tabular}{cccccccccc}
        \toprule
         & \multicolumn{3}{c}{Composition} & \multicolumn{2}{c}{Train} & \multicolumn{2}{c}{Val} & \multicolumn{2}{c}{Test}  \\
         \cmidrule(lr){2-4} \cmidrule(lr){5-6} \cmidrule(lr){7-8} \cmidrule(lr){9-10}
         Datasets & $|\mathcal{A}|$ & $|\mathcal{O}|$ & $|\mathcal{A}|\times|\mathcal{O}|$ & $|\mathcal{C}_{s}|$ & $|\mathcal{X}|$ & $|\mathcal{C}_{s}|$ / $|\mathcal{C}_{u}|$ & $|\mathcal{X}|$ & $|\mathcal{C}_{s}|$ / $|\mathcal{C}_{u}|$ & $|\mathcal{X}|$
         \\ \midrule
        Clothing16K~\cite{zhang2022learning} & 9 & 8 & 72 & 18 & 7242 & 10 / 10 & 5515 & 9 / 8 & 3413\\
        UT-Zappos50K~\cite{yu2014fine} & 16 & 12 & 192 & 83 & 22998 & 15 / 15 & 3214 & 18 / 18 & 2914 \\
        C-GQA~\cite{naeem2021learning} & 413 & 674 & 278362 & 5592 & 26920 & 1252 / 1040 & 7280 & 888 / 923 & 5098 \\
        \bottomrule
    \end{tabular}}
    \caption{Summary of data split statistics.}
    \label{tab:data-splits}
    \vspace{-10pt}
\end{table}

\paragraph{Open-world setting} In addition to the standard closed-world setting, we also evaluate our model on the open-world setting~\cite{mancini2021open}, which is neglected by most previous works. The open-world setting considers all possible compositions, which requires a much larger testing space than the closed-world setting during inference. Taking UT-Zappos50K as an example (see~\cref{tab:data-splits}), the closed world only considers 36 compositions in the testing set while the open world considers total 192 compositions, in which $\sim$81\% are ignored under the standard closed-world setting.

\paragraph{Evaluation metrics} Since CZSL models have an inherent bias for seen compositions, we follow the generalized CZSL evaluation protocol~\cite{purushwalkam2019task}. To overcome the negative bias for seen compositions, we apply different calibration terms to unseen compositions and compute the corresponding top-1 accuracy of seen and unseen compositions, where a larger bias makes higher unseen accuracy and lower seen accuracy, and vice versa.  We treat seen accuracy as $x$-axis and unseen accuracy as $y$-axis to derive an unseen-seen accuracy curve. We can then compute the area under curve (AUC), the best harmonic mean, the best seen accuracy, and the best unseen accuracy from the curve. In our experiments, we report these four metrics for evaluation, among which AUC is the most representative and stable metric for measuring CZSL model performance. \hsz{Note that the attribute accuracy or the object accuracy alone does not reflect CZSL performance, because the individual accuracy on attribute or object does not necessarily decide the accuracy of their composition.}

\paragraph{Implementation details}
We use a frozen ViT-B-16~\cite{dosovitskiy2020vit} backbone pretrained with DINO~\cite{caron2021emerging} on ImageNet~\cite{deng2009imagenet} in a self-supervised manner as our visual feature extractor. The ViT-B-16 outputs contain 197 tokens (1 \texttt{[CLS]} and 196 patch tokens) of 768 dimensions. For three attention disentangler modules, we implement one-layer multi-head attention framework following~\cite{vaswani2017attention}, changing the single input to paired inputs for cross-attentions. The embedders $\pi_a$, $\pi_c$, $\pi_o$ are the two-layer MLPs following the previous works~\cite{mancini2021open, zhang2022learning}, projecting the 768-dimension visual features to 300-dimension word embedding space. The word embedding prototypes are initialized with word2vec~\cite{mikolov2013distributed} for all datasets and learnable during training. The composition function $\psi$ is one linear layer. We train our model using Adam optimizer~\cite{kingma2015adam} with a learning rate of $5\times 10^{-6}$ for UT-Zappos50K and Clothing16K, and $5\times 10^{-5}$ for C-GQA. All models are trained with 128 batch size for 300 epochs.

\begin{table*}[t]
    \centering
    \scalebox{0.75}{
    \begin{tabular}{l>{\columncolor{tabcolor}}cccccc>{\columncolor{tabcolor}}cccccc>{\columncolor{tabcolor}}cccccc}
        \toprule
         Closed-world & \multicolumn{6}{c}{Clothing16K} & \multicolumn{6}{c}{UT-Zappos50K} & \multicolumn{6}{c}{C-GQA} \\
         \cmidrule(lr){2-7} \cmidrule(lr){8-13} \cmidrule(lr){14-19}
         Models & AUC & HM & Seen & Unseen & Attr & Obj & AUC & HM & Seen & Unseen & Attr & Obj & AUC & HM & Seen & Unseen & Attr & Obj \\
         \midrule
         SymNet~\cite{li2020symmetry} & 78.8 & 79.3 & 98.0 & 85.1 & 75.6 & 84.1 & 32.6 & 45.6 & 60.6 & 68.6 & 48.2 & 77.0 & 3.1 & 13.5 & 30.9 & 13.3 & 11.4 & 34.6 \\
         CompCos~\cite{mancini2021open} & 90.3 & 87.2 & 98.5 & 96.8 & \textbf{90.2} & 91.8 & 31.8 & 48.1 & 58.8 & 63.8 & 45.5 & 72.4 & 2.9 & 12.8 & 30.7 & 12.2 & 10.4 & 33.9 \\
         GraphEmb~\cite{naeem2021learning} & 89.2 & 84.2 & 98.0 & 97.4 & 90.0 & 93.1 & 34.5 & 48.5 & 61.6 & \textbf{70.0} & \textbf{50.8} & \textbf{77.1} & 3.8 & 15.0 & 32.3 & 14.9 & 13.8 & 33.2 \\
         Co-CGE~\cite{mancini2022learning} & 88.3 & 87.9 & 98.5 & 94.7 & 87.4 & 91.4 & 30.8 & 44.6 & 60.9 & 62.6 & 46.0 & 73.5 & 3.6 & 14.7 & 31.6 & 14.3 & 12.6 & 34.6 \\
         SCEN~\cite{li2022siamese} & 78.8 & 78.5 & 98.0 & 89.6 & 81.2 & 85.4 & 30.9 & 46.7 & \textbf{65.7} & 62.9 & 44.0 & 74.4 & 3.5 & 14.6 & 31.7 & 13.4 & 10.7 & 31.4 \\ 
         IVR~\cite{zhang2022learning} & 90.6 & 86.6 & \textbf{99.0} & 97.0 & 89.3 & \textbf{93.6} & 34.3 & 49.2 & 61.5 & 68.1 & 48.4 & 74.6 & 2.2 & 10.9 & 27.3 & 10.0 & 10.3 & \textbf{37.5} \\
         OADis~\cite{Saini_2022_CVPR} & 88.4 & 86.1 & 97.7 & 94.2 & 84.9 & 93.1 & 32.6 & 46.9 & 60.7 & 68.8 & 49.3 & 76.9 & 3.8 & 14.7 & 33.4 & 14.3 & 8.9 & 36.3 \\
         \midrule
         \framework (ours) & \textbf{92.4} & \textbf{88.7} & 98.2 & \textbf{97.7} & \textbf{90.2} & \textbf{93.6} & \textbf{35.1} & \textbf{51.1} & 63.0 & 64.3 & 46.3 & 74.0 & \textbf{5.2} & \textbf{18.0} & \textbf{35.0} & \textbf{17.7} & \textbf{16.8} & 32.3\\ 
        \bottomrule
    \end{tabular}}
    \caption{Closed-world results on three datasets. We report the area under curve (AUC), the best harmonic mean (HM), the best seen accuracy (Seen), the best unseen accuracy (Unseen), the attribute accuracy (Attr), and the object accuracy (Obj) of the unseen-seen accuracy curve under the closed world-setting. AUC is the core CZSL metric. All models use the same DINO ViT-B-16 backbone.} 
    \label{tab:cw-results}
\end{table*}

\begin{table*}[t]
    \centering
    \scalebox{0.75}{
    \begin{tabular}{l>{\columncolor{tabcolor}}cccccc>{\columncolor{tabcolor}}cccccc>{\columncolor{tabcolor}}cccccc}
        \toprule
         Open-world & \multicolumn{6}{c}{Clothing16K} & \multicolumn{6}{c}{UT-Zappos50K} & \multicolumn{6}{c}{C-GQA} \\
         \cmidrule(lr){2-7} \cmidrule(lr){8-13} \cmidrule(lr){14-19}
         Models & AUC & HM & Seen & Unseen & Attr & Obj  & AUC & HM & Seen & Unseen & Attr & Obj & AUC & HM & Seen & Unseen & Attr & Obj  \\
         \midrule
         SymNet~\cite{li2020symmetry} & 57.4 & 68.3 & 98.2 & 60.7 & 57.6 & 81.2 & 25.0 & 40.6 & 60.4 & 51.0 & 38.2 & \textbf{75.0} & 0.77 & 4.9 & 30.1 & 3.2 & 18.4 & 37.5 \\
         CompCos~\cite{mancini2021open} & 64.1 & 70.8 & 98.2 & 69.8 & 71.7 & 83.7 & 20.7 & 36.0 & 58.1 & 46.0 & 36.4 & 71.1 & 0.72 & 4.3 & 32.8 & 2.8 & 15.1 & 37.8 \\
         GraphEmb~\cite{naeem2021learning} & 62.0 & 68.3 & 98.5 & 69.7 & 71.8 & 82.4 & 23.5 & 40.0 & 60.6 & 47.0 & 37.1 & 69.3 & 0.81 & 4.8 & 32.7 & 3.2 & 17.2 & 36.7 \\
         Co-CGE~\cite{mancini2022learning} & 59.3 & 69.2 & 98.7 & 63.8 & 68.5 & 76.2 & 22.0 & 40.3 & 57.7 & 43.4 & 33.9 & 67.2 & 0.48 & 3.3 & 31.1 & 2.1 & 15.5 & 35.7 \\
         SCEN~\cite{li2022siamese}& 53.7 & 61.5 & 96.7 & 62.3 & 63.6 & 79.1 & 22.5 & 38.0 & \textbf{64.8} & 47.5 & 34.9 & 73.3 & 0.34 & 2.5 & 29.5 & 1.5 & 14.8 & 32.3 \\ 
         IVR~\cite{zhang2022learning} & 63.6 & 72.0 & 98.7 & 69.0 & 70.3 & 84.8 & 25.3 & 42.3 & 60.7 & 50.0 & 38.4 & 71.4 & 0.94 & 5.7 & 30.6 & 4.0 & 16.9 & 36.5 \\
         OADis~\cite{Saini_2022_CVPR} & 53.4 & 63.2 & 98.0 & 58.6 & 57.3 & \textbf{85.4} & 25.3 & 41.6 & 58.7 & \textbf{53.9} & \textbf{40.3} & 74.7 & 0.71 & 4.2 & 33.0 & 2.6 & 14.6 & \textbf{39.7} \\
         \midrule
         \framework (ours) & \textbf{68.0} & \textbf{74.2} & \textbf{99.0} & \textbf{73.1} & \textbf{75.0} & 84.5 & \textbf{27.1} & \textbf{44.8} & 62.4 & 50.7 & 39.9 & 71.4 & \textbf{1.42} & \textbf{7.6} & \textbf{35.1} & \textbf{4.8} & \textbf{22.4} & 35.6 \\ 
        \bottomrule
    \end{tabular}}
    \caption{Open-world results on three datasets. Different from~\cref{tab:cw-results}, open-world setting considers all possible compositions in testing.} 
    \label{tab:ow-results}
\end{table*}

\subsection{Comparison}
\hsznew{To ensure a fair comparison and demonstrate that our improvement over baseline models is not merely by ViT, we adopt ViT backbone to state-of-the-art CZSL models and \emph{re-train} all models.} We compare our method with them: \hsznew{(1)~OADis~\cite{Saini_2022_CVPR} disentangles attribute and object features from spatial convolutional maps;} (2)~SymNet~\cite{li2020symmetry} introduces the symmetry principle of attribute-object transformation and group theory as training objectives; (3)~CompCos~\cite{mancini2021open} extends CZSL to an open-world setting considering all possible compositions during inference, proposing a feasibility score based on data statistics to remove unfeasible compositions; (4)~GraphEmb~\cite{naeem2021learning} and Co-CGE~\cite{mancini2022learning} propose to use graph convolutional networks (GCN) to represent attribute-object relationships and compositions; (5)~SCEN~\cite{li2022siamese} projects visual features to a Siamese contrastive space to capture concept prototypes, and introduces complex state transition module to produce virtual compositions; (6)~IVR~\cite{zhang2022learning} proposes to disentangle visual features into concept-invariant domains from a perspective of domain generalization, by masking specific channels of visual features. 

\paragraph{Closed-world evaluation} In~\cref{tab:cw-results}, we compare our \framework model with the state-of-the-art methods. \framework consistently outperforms others by a significant margin. \framework increases the core metric AUC by 1.8 on Clothing16K, 0.6 on UT-Zappos50K, and 1.4 on C-GQA ($\sim$37\% relatively). Similarly, \framework increases the best harmonic mean (HM) by 0.8\% on Clothing16K, 1.9\% on UT-Zappos50K, and 3.0\% on C-GQA. We notice that SymNet~\cite{li2020symmetry} and SCEN~\cite{li2022siamese} perform badly on Clothing16K. The reason might be that not learning concept prototypes harms the word embedding expressivity on small-scale concepts. We also notice that IVR~\cite{zhang2022learning} performs very well on curated datasets Clothing16K and UT-Zappos50K but badly on larger-scale real-world dataset C-GQA. We hypothesize ideal concept-invariant domains might be difficult to learn from natural images and large-scale concepts of C-GQA. In contrast, our \framework model achieves state-of-the-art performance on all datasets.

\paragraph{Open-world evaluation} In~\cref{tab:ow-results}, we consider the open-world setting to compare our \framework with other methods. Likewise, \framework also performs the best among all methods under open-world setting. \framework increases AUC by 3.9 on Clothing16K, 1.8 on UT-Zappos50K, and 0.48 on C-GQA ($\sim$51\% relatively). \framework also increases the best harmonic mean (HM) by 2.2\% on Clothing16K, 2.5\% on UT-Zappos50K, and 1.9\% on C-GQA ($\sim$33\% relatively). From the above results, \framework surpasses others by a larger margin on open-world AUC and HM than closed-world ones, indicating \framework maintains utmost efficiency when turning from the closed world to the open world. It is worth mentioning that \framework does not apply any special operations (\eg, feasibility masking~\cite{mancini2021open}) for the open world and deals with the two settings in exactly the same way. 
IVR~\cite{zhang2022learning} keeps its performance to a great extent but still lags behind our method significantly.

\begin{table*}[t]
\begin{minipage}[t]{0.44\linewidth}
    \centering
    \scalebox{0.72}{
    \begin{tabular}{lcccc>{\columncolor{tabcolor}}cccc}
        \toprule
         & CA & AA & OA & Reg & AUC & HM & Seen & Unseen\\
         \midrule
         (0) & \xmark & \xmark & \xmark & \xmark & 23.8 & 41.1 & 59.0 & 48.9 \\
         (1) & self & \xmark & \xmark & \xmark & 25.3 & 42.3 & 61.1 & 49.9 \\
         (2) & self & self & self & \xmark & 26.7 & 44.6 & 61.9 & 49.8 \\
         (3) & self & cross & cross & \xmark & 26.9 & 44.5 & \textbf{63.4} &48.7 \\
         (4) & self & cross & cross & \cmark &  \textbf{27.1} & \textbf{44.8} & 62.4 & \textbf{50.7}\\
         
        \bottomrule
    \end{tabular}}
    \caption{Ablate the components in \framework on open-world UT-Zappos50K. CA, AA, and OA denote composition, attribute, and object attention. Reg denotes the regularization term. We test self- or cross-attention for AA and OA.} 
    \label{tab:model-ab}
\end{minipage}
\hspace{2mm}
\begin{minipage}[t]{0.54\textwidth}
\centering
    \scalebox{0.68}{
    \begin{tabular}{ll>{\columncolor{tabcolor}}cccc>{\columncolor{tabcolor}}cccc}
        \toprule
        & & \multicolumn{4}{c}{C-GQA} & \multicolumn{4}{c}{Clothing16K} \\
        \cmidrule(lr){3-6} \cmidrule(lr){7-10}
        & Inference formulation  & AUC & HM & Seen & Unseen & AUC & HM & Seen & Unseen\\
        \midrule
        (0) & $p(c)$ & 4.6 & 16.8 & \textbf{35.1} & 16.0 & \textbf{92.4} & \textbf{88.8} & \textbf{98.2} & \textbf{97.7} \\
        (1) & $p(a) \cdot p(o)$ &  4.0 & 15.8 & 31.4 & 15.1 & 57.3 & 66.3 & 96.7 & 63.1\\
        (2) & $p(c) + p(a) \cdot p(o)$ & \textbf{5.2} & \textbf{18.0} & 35.0 & \textbf{17.7} & 90.4 & 85.9 & 98.2 & 97.0\\
        (3) & $p(c) + \beta \cdot p(a) \cdot p(o)$ & \textbf{5.2} & \textbf{18.0} & 35.0 & \textbf{17.7} & \textbf{92.4} & 88.7 & \textbf{98.2} & \textbf{97.7} \\
        \bottomrule
    \end{tabular}}
    \caption{Results on closed-world Clothing16K and C-GQA using different inference formulations. Rows (0)-(2) respectively represents the cases when $\beta=0.0$, $\beta=+\infty$, and $\beta=1.0$. Row (3) is our inference formulation, which applies an $\beta$ optimized on the validation set.} 
    \label{tab:eval-ab}
\end{minipage}
\end{table*}

\subsection{Ablation study}
\paragraph{Backbone: ResNet \textit{vs} ViT}
\hsznew{Our work leverages ViT as the default backbone to excavate more high-level sub-space information, while ResNet18 is the most common backbone in previous works. 
In~\Cref{tab:backbone}, we compare our \framework to OADis~\cite{Saini_2022_CVPR}  with both backbones. Our \framework performs similarly to OADis with ResNet18, but outperforms it significantly with ViT. Additionally, we present an ablation study on different components of our method with the ResNet18 backbone in the Appendix. These experiments indicate that our model benefits from ViT and all components of our method are effective regardless of the backbone.}

\begin{table}[h]
    \centering
    \scalebox{0.75}{
    \begin{tabular}{ll>{\columncolor{tabcolor}}cc>{\columncolor{tabcolor}}cc>{\columncolor{tabcolor}}cc}
        \toprule
          \multicolumn{2}{l}{Closed-world} & \multicolumn{2}{c}{Clothing16K} & \multicolumn{2}{c}{UT-Zappos50K} & \multicolumn{2}{c}{C-GQA} \\
         \cmidrule(lr){3-4} \cmidrule(lr){5-6} \cmidrule(lr){7-8}
         Backbone & Models & AUC & HM & AUC & HM & AUC & HM \\
         \midrule
         \multirow{2}{*}{ResNet18} & OADis~\cite{Saini_2022_CVPR} & 85.5 & 84.7 & \textbf{30.0} & 44.4 & \textbf{3.1} & 13.6 \\
          & \framework (ours) & \textbf{87.2} & \textbf{85.1}  & 29.5 & \textbf{47.0} & \textbf{3.1} & \textbf{13.7} \\
         \midrule
         \multirow{2}{*}{ViT-B-16} & OADis~\cite{Saini_2022_CVPR} & 88.4 & 86.1  & 32.6 & 46.9 & 3.8 & 14.7 \\
          & \framework (ours) & \textbf{92.4} & \textbf{88.7}  & \textbf{35.1} & \textbf{51.1} & \textbf{5.2} & \textbf{18.0} \\
        \bottomrule
    \end{tabular}}
    \caption{Compare \framework and OADis~\cite{Saini_2022_CVPR} with ResNet18 and ViT.} 
    \label{tab:backbone}
    \vspace{-5pt}
\end{table}

\paragraph{Different parts of \framework} We evaluate the effectiveness of attention disentanglers (composition, attribute, and object attention) and the regularization term in our model. We report the ablation study results on the open-world UT-Zappos50K in~\cref{tab:model-ab}. Rows~(0)-(2) show attention disentanglers can significantly improve the performance. Rows~(2)-(3) show that cross-attention learns disentangled concepts better than self-attention for AA and OA. Rows~(3)-(4) show the regularization term can further benefit the visual disentanglement, improving the unseen accuracy and overall AUC.

\paragraph{Inference formulation} We also investigate the effect of our inference formulation $p(c) + \beta \cdot p(a) \cdot p(o)$ in~\cref{tab:eval-ab}. We report the results with extreme values of $\beta$, \ie, $\beta=0.0$ and $\beta=1.0$. Note that $\beta=0.0$ means only using composition probability for prediction. In addition, we also test the performance only using the product of attribute and object probabilities $p(a) \cdot p(o)$. We can observe that the best fixed $\beta$ value is unfixed among datasets. For example, $\beta=1.0$ gives the highest AUC for C-GQA in row (2) while $\beta=0.0$ for Clothing16K in row (0). In contrast, our validated $\beta$ consistently gives the best inference results for both datasets. Another observation on C-GQA is that $p(a) \cdot p(o)$ alone is not a good prediction, but adding it to $p(c)$ can increase the unseen accuracy. This indicates that the disentangled attribute prediction $p(a)$ and object prediction $p(o)$ indeed enhance the unseen generalization for CZSL problem.

\paragraph{Effect of regularization term} We propose an EMD-adapted regularization term at the attention level to force attentions to disentangle the concept of interest. We also investigate the effect of applying the regularization term at the feature level. Specifically, 
we compare our EMD-based distance to the cosine and euclidean feature distances. The results on open-world UT-Zappos50K are shown in~\cref{tab:reg-ab}. Our EMD-based regularization outperforms other distance forms, because our attention-level EMD distance considers token-wise similarity capturing the specific concept-related attention responses.
\begin{table}[h]
   \centering
   \setlength{\tabcolsep}{20pt}
    \scalebox{0.7}{
    \begin{tabular}{l>{\columncolor{tabcolor}}cccc}
        \toprule
        Reg & AUC & HM & Seen & Unseen \\
        \midrule
        Cosine & 26.8 & 44.7 & \textbf{63.0} & 48.6 \\
        Euclidean & 26.2 & 44.3 & 62.6 & 47.5 \\
        Ours (EMD) & \textbf{27.1} & \textbf{44.8} & 62.4 & \textbf{50.7}\\
        \bottomrule
    \end{tabular}}
    \caption{Comparison of different regularization terms on open-world UT-Zappos50K.} 
    \label{tab:reg-ab} 
    \vspace{-5pt}
\end{table}

\subsection{Qualitative analysis}
Visual disentanglement in feature space is hard to visualize~\cite{Saini_2022_CVPR}. Inspired by previous work attempts~\cite{Saini_2022_CVPR,zhang2022learning,li2020symmetry,nagarajan2018attributes}, we conduct qualitative analysis of image and text retrieval to show how our \framework model correlates the visual image and the concept composition. In addition, to further validate \framework is efficient to disentangle visual concepts, we conduct unseen-to-seen image retrieval based on their visual concept features extracted by attribute and object attentions.

\begin{figure*}[t]
     \centering
     \begin{subfigure}[b]{0.32\textwidth}
         \centering
         \includegraphics[width=0.92\linewidth]{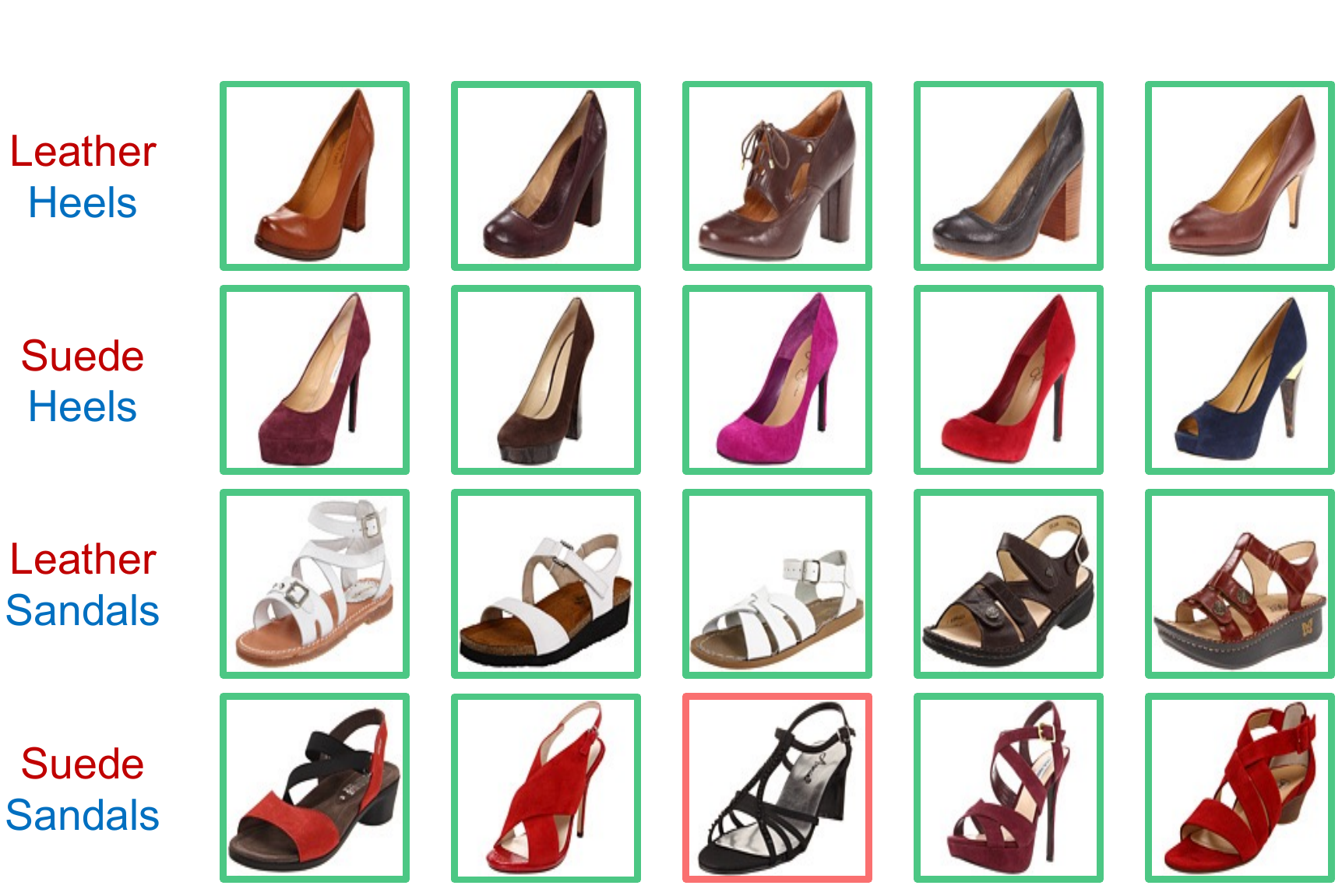}
    \caption{Top-5 text-to-image retrieval.}
    \label{fig:wrd2img}
     \end{subfigure}
     \hfill
     \begin{subfigure}[b]{0.35\textwidth}
    \centering
    \includegraphics[width=\linewidth]{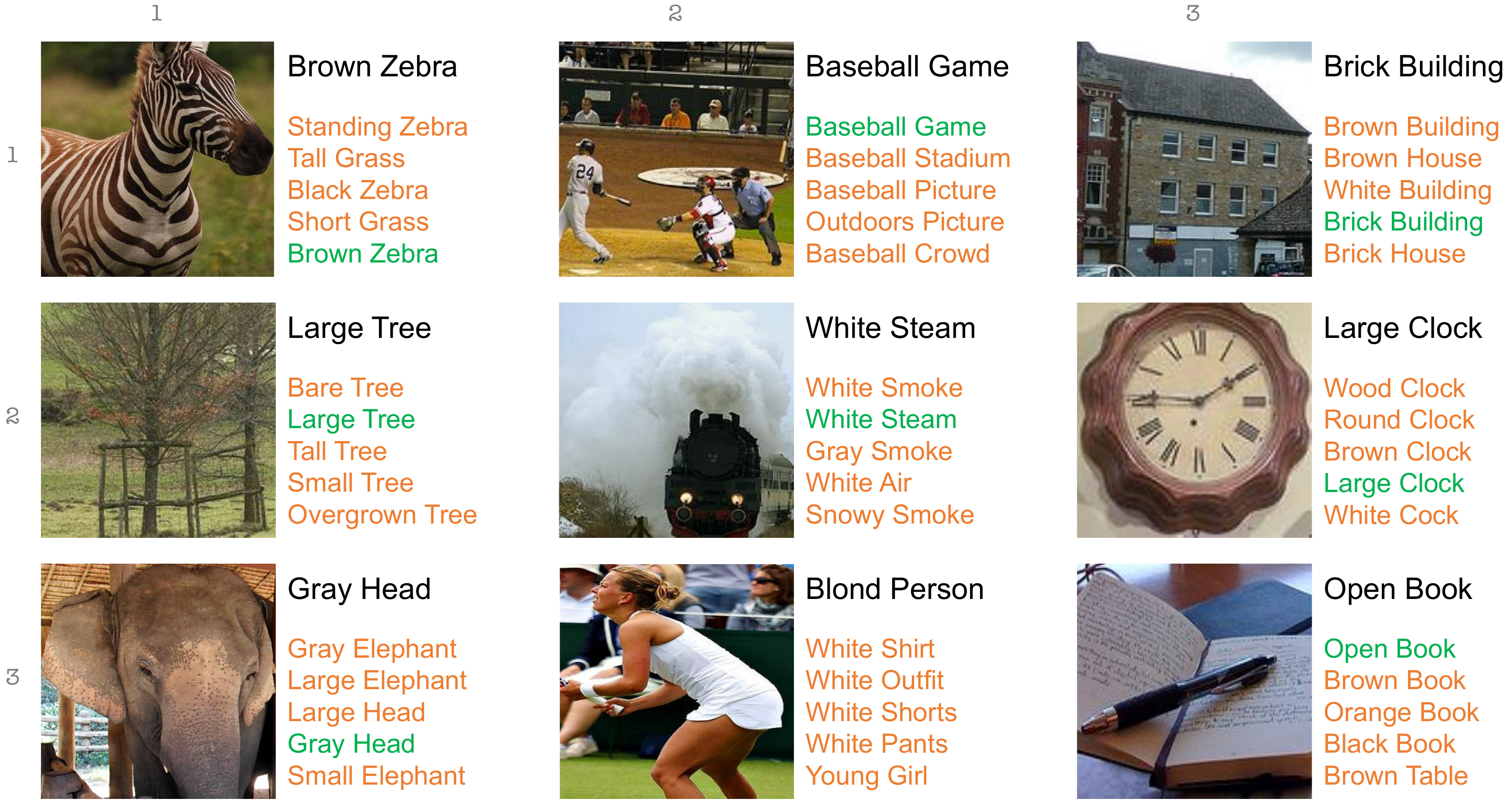}
    \caption{Top-5 image-to-text retrieval.}
    \label{fig:img2wrd}
     \end{subfigure}
     \hfill
     \begin{subfigure}[b]{0.32\textwidth}
    \centering
    \includegraphics[width=0.9\linewidth]{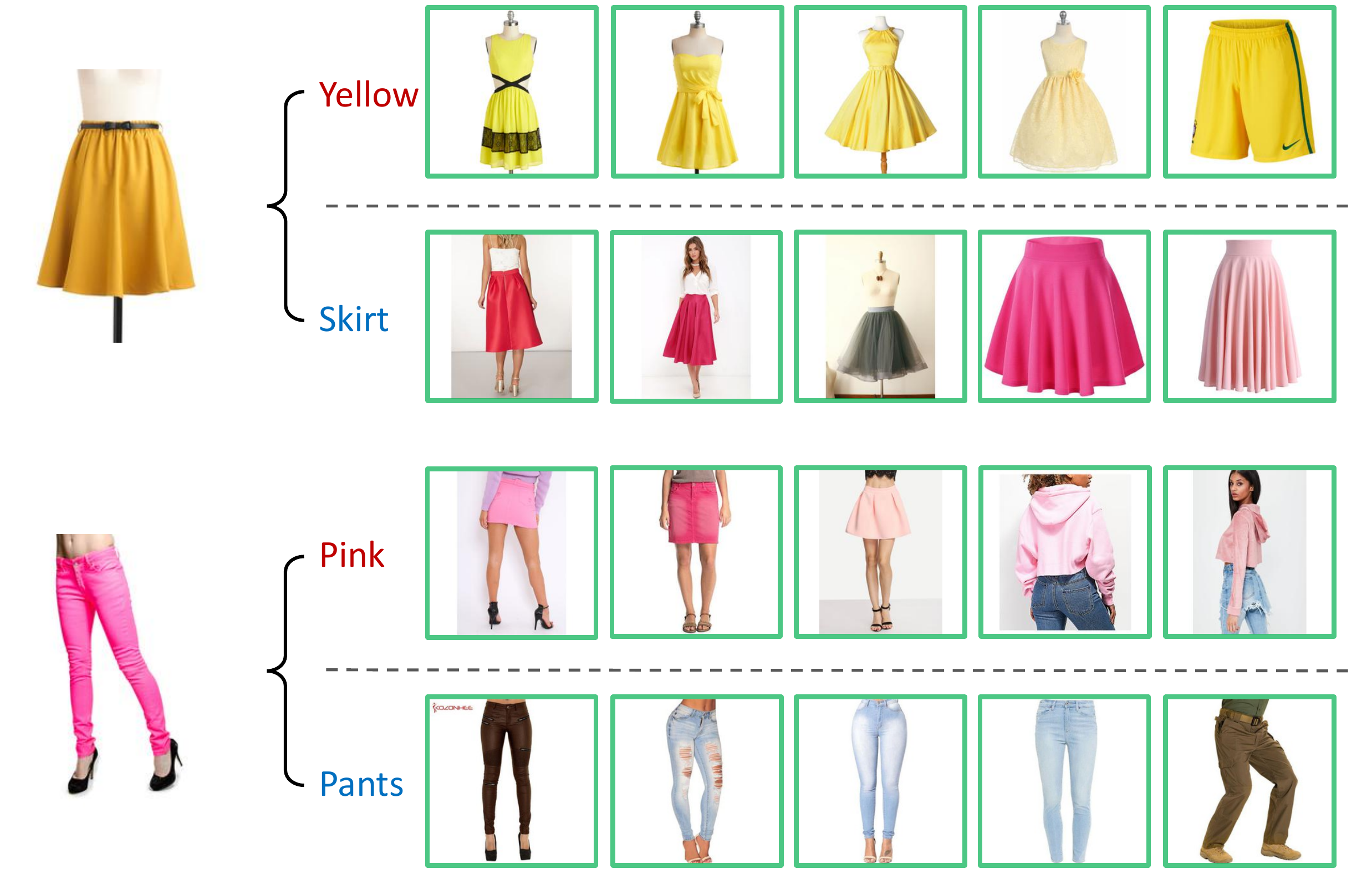}
    \caption{Top-5 visual concept retrieval.}
    \label{fig:concept-retrieve}
     \end{subfigure}
    \caption{\hsznew{Qualitative analysis. (a) In the last row of ``suede sandals", the wrong image (red box) is ``fake leather sandals". (b) Each image has the ground-truth label (black text) and 5 retrieval results (colored text), in which the green text is the correct prediction. (c) We retrieve images sharing the same visual concepts by their visual concept features for unseen images of ``yellow skirt" and ``pink pants".}}
    \label{fig:qualitative}
    \vspace{-4pt}
\end{figure*}

\paragraph{Image and text retrieval}
We first consider text-to-image retrieval. Given a text composition, \eg, ``leather heels", we embed it and retrieve the top-5 closest visual features based on the feature distance. We display four text compositions of the different objects sharing the same attributes and vice versa in~\cref{fig:wrd2img}. We can observe that the retrieved images are correct in most of the cases. One exception is when retrieving ``suede sandals", the third closest image is ``fake leather sandals". Although ``suede sandals" and ``fake leather sandals" are not the same composition, they are quite visually similar. We then consider image-to-text retrieval, shown in~\cref{fig:img2wrd}. Given an image, \eg, the image of a ``brown zebra", we extract its visual feature and retrieve the top-5 closest text composition embeddings. It is difficult to retrieve the ground-truth label in the top-1 closest text composition, but all top-5 results are all semantically related to the image. We take the image of ``blond person" (row 3, col 2) as an example. Although the text composition ``blond person" is not retrieved in the top-5 matches, the retrieved results ``white shirts", ``white outfit", ``white shorts", ``white pants", and ``young girl" are all reasonable and actually present in the image. Image and text retrieval experiments validate that our \framework efficiently projects visual features and word embeddings into a uniform space.
\vspace{-2pt}

\paragraph{Visual concept retrieval}
Because the attribute and the object are visually coupled in an image to a high degree of entanglement, it is challenging to visualize the disentanglement in feature space~\cite{Saini_2022_CVPR}. Saini \etal~\cite{Saini_2022_CVPR} retrieve single attribute or object text from test images. However, this process is the same as multi-label classification and insufficient to validate that disentangled visual concepts are learned from images. \hsz{Based on the disentanglement ability of ADE, we construct a visual concept retrieval experiment to investigate the distances between visual concept features, \ie, the embedded attribute feature $\pi_a(v_a)$ and the embedded object feature $\pi_o(v_o)$, extracted from different images. Prior to our work, no existing models can do so, because none of them extracts concept-exclusive features like ADE.} The results are shown in~\cref{fig:concept-retrieve}. We first extract attribute features and object features from all seen images. Given an unseen image, we retrieve the top-5 closest images by measuring the feature distance between the attribute feature of the given image and that of all seen images, and the same goes for the object feature. For the image of ``yellow skirt", all retrieval results for the visual concept ``yellow" are all  ``yellow \texttt{[OBJ]}", and all retrieval results for ``skirt" are ``\texttt{[ATTR]} skirt". For the ``pink pants" image, our model also perfectly retrieves the visual concepts, \ie, the attribute ``pink" and the object ``pants". Our experimental results demonstrate that our \framework model is effective to disentangle visual concepts from seen compositions and combine learned concept knowledge into unseen compositions.

%% file: 10_conclusion.tex
\vspace{-5pt}
\section{Conclusion}
\label{sec:conclusion}
In this paper, we tackle the problem of compositional zero-shot learning (CZSL). To disentangle visual concepts from the attribute-object composition, we propose \framework adopting cross-attentions to learn the individual concept from paired concept-sharing images. To constrain the disentanglers to learn the concept of interest, we employ a regularization term adapted from the earth mover's distance (EMD), which is used as a feature similarity metric in the cross-attention module. Moreover, to exploit the attribute and object prediction ability of \framework, we improve the inference process by combining attribute, object, and composition probabilities into the final prediction score. We empirically demonstrate \framework outperforms the current state-of-the-art methods under both closed- and open-world settings. We also conduct a comprehensive qualitative analysis to validate the disentanglement ability of attention disentanglers in \framework. 
\paragraph{Limitations} \hsznew{Like existing CZSL methods, it is time- and computation-consuming to derive all composition embeddings when the numbers of attributes and objects are large. 
Moreover, it remains an open challenge to exploit concepts based on the actual semantics rather than solely on text; for instance, the “open” attribute in “open curtain” and “open computer” has completely different meanings.} 
\paragraph{Acknowledgements} 
This work is partially supported by Hong Kong Research
Grant Council - Early Career Scheme (Grant No. 27208022) and HKU Seed Fund for Basic Research.

%% file: 12_appendix.tex
\appendix
\label{sec:appendix}

\section{Coefficient $\beta$ during inference}
At inference time, we use a coefficient $\beta$ to strike a balance between the composition score $p(c)$ and the product of the attribute and object scores $p(a)\cdot p(o)$. The final prediction score $\tilde{p}(c)$ is given by:
\begin{equation}
\setlength{\abovedisplayskip}{5pt}
\setlength{\belowdisplayskip}{5pt}
    \tilde{p}(c) =  p(c) + \beta \cdot p(a) \cdot p(o)
    \label{eq:supp-predict}
\end{equation}
where $c \in \mathcal{C}_{test}$ and $c=(a, o) \in \mathcal{A} \times \mathcal{O}$. We first fix $\beta=1.0$ during training, and then validate $\beta=0.0, 0.1, \cdots, 1.0$ to choose the best $\hat{\beta}$ on the validation set. We report the chosen $\hat{\beta}$ values for different datasets in~\cref{tab:beta}.

\begin{table}[h]
    \centering
    \setlength{\tabcolsep}{15pt}
    \scalebox{0.82}{
    \begin{tabular}{lcc}
    \toprule
    Datasets & Closed-world & Open-world \\
    \midrule
    Clothing16K~\cite{zhang2022learning} & $\hat{\beta}=0.1$ & $\hat{\beta}=0.1$ \\
    UT-Zappos50K~\cite{yu2014fine} & $\hat{\beta}=0.9$ & $\hat{\beta}=0.9$ \\
    C-GQA~\cite{naeem2021learning} & $\hat{\beta}=1.0$ & $\hat{\beta}=0.7$ \\
    \bottomrule
    \end{tabular}}
    \caption{The chosen $\hat{\beta}$ values for different datasets under the closed-world and the open-world settings.}
    \label{tab:beta}
\end{table}

\section{Unseen-seen accuracy curve}
For the CZSL evaluation metric, we follow the generalized evaluation protocol~\cite{purushwalkam2019task, chao2016empirical}. To overcome the negative bias on seen compositions, we use a calibration term for unseen compositions. This calibration term increases unseen composition scores and leads to the following classification rule:
\begin{equation}
\setlength{\abovedisplayskip}{5pt}
\setlength{\belowdisplayskip}{5pt}
    \hat{c} = \mathop{\arg\max}_{c \in \mathcal{C}_{test}} \ \tilde{p}(c) + \gamma\mathbb{I}[c \in \mathcal{C}_{u}]
\end{equation}
where the prediction $\tilde{p}(c)$ is computed by~\cref{eq:supp-predict}, $\gamma$ is the calibration term, $\mathbb{I}[\cdot] \in \{0, 1\}$ indicates whether or not $c$ is an unseen composition, \ie, $c \in \mathcal{C}_{u}$. When using different calibration terms, we can obtain different paired top-1 accuracy of seen and unseen compositions. 
Without any constraints, we can obtain the highest unseen accuracy by  $\gamma=+\infty$  and the highest seen accuracy by $\gamma=-\infty$, leading to trivial solutions.
To construct a feasible list of different calibration values, we first compute $\gamma_i$ for each image $i$ of unseen compositions:
\begin{equation}
\setlength{\abovedisplayskip}{5pt}
\setlength{\belowdisplayskip}{5pt}
    \gamma_i = \mathop{\max}_{c \in \mathcal{C}_{s}}\ \tilde{p}(c\ |\ i) - \tilde{p}(c_{i}\ |\ i)
    \label{eq:calibration}
\end{equation}
where $c_i \in \mathcal{C}_{u}$ is the ground-truth composition of the image $i$ and $\tilde{p}(c\ |\ i)$ denotes the prediction score of composition $c$ for the image $i$. A list of $\gamma_i$ can be derived by applying~\cref{eq:calibration} on all unseen-composition images. We then sort the list, in which the smallest value makes the highest seen accuracy and the largest value makes the highest unseen accuracy. We pick $\gamma_i$ in the list with a specific interval and obtain multiple seen-unseen accuracy pairs. In this way, we can plot a curve with all scatters of seen and unseen accuracy, from which the evaluation metrics AUC (area under curve) and HM (the best harmonic mean accuracy) are obtained. 

\begin{figure}[t]
    \centering
    \includegraphics[width=\linewidth]{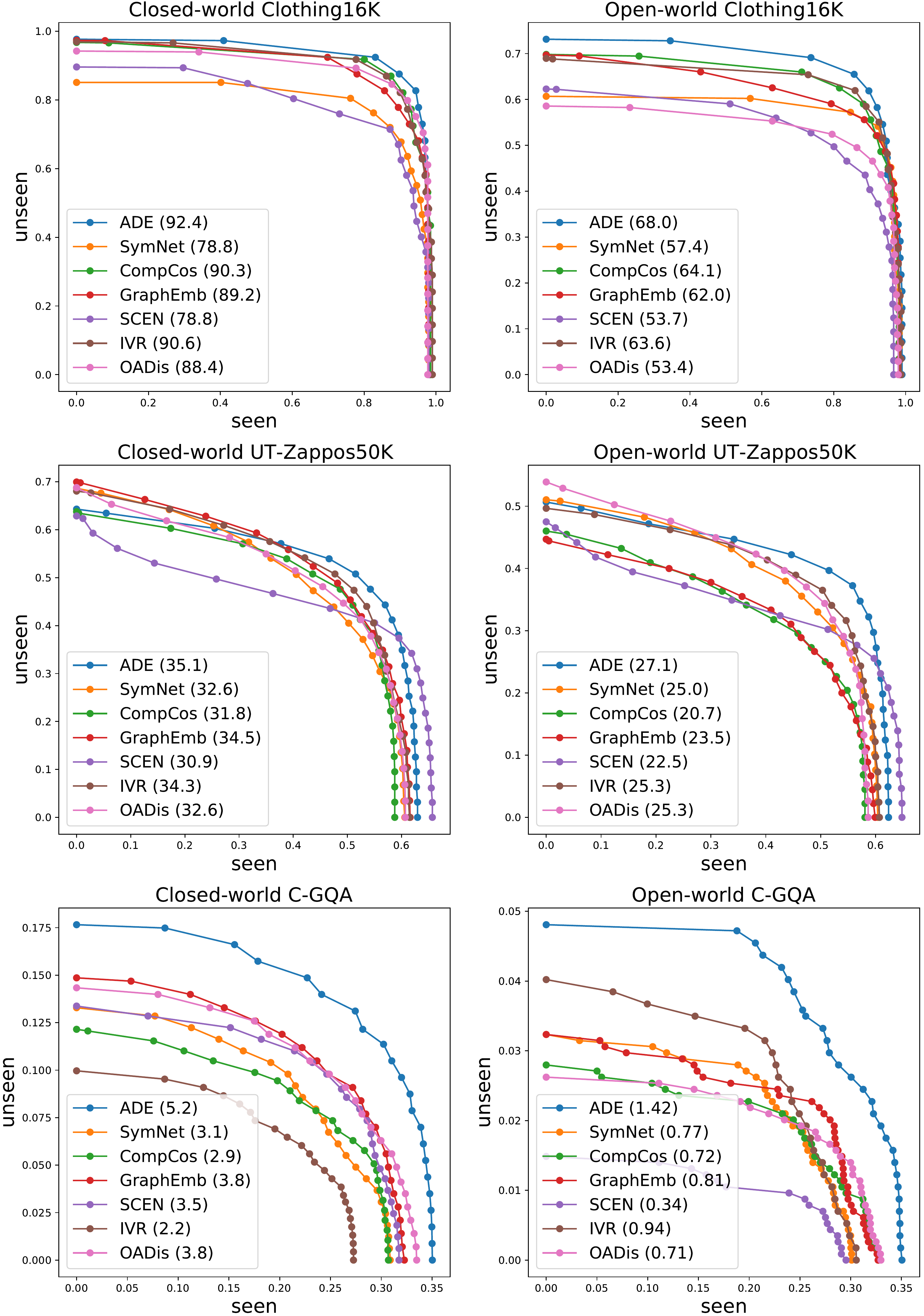}
    \caption{Unseen-seen accuracy curve on Clothing16K~\cite{zhang2022learning}, UT-Zappos50K~\cite{yu2014fine}, and C-GQA~\cite{naeem2021learning} under the closed-world and open-world settings. We compare our \framework with SymNet~\cite{li2020symmetry}, CompCos~\cite{mancini2021open}, GraphEmb~\cite{naeem2021learning}, SCEN~\cite{li2022siamese}, IVR~\cite{zhang2022learning}, and OADis~\cite{Saini_2022_CVPR}. Area under curve (AUC) is reported in brackets.}
    \label{fig:acc-curve}
\end{figure}

In~\cref{fig:acc-curve}, we show the unseen-seen accuracy curve of all compared CZSL methods on all datasets under the closed-world and open-world settings. With the increase of the calibration value, the classification accuracy of seen compositions decreases while the accuracy of unseen compositions increases. The evaluation metrics in the paper, \ie, area under curve (AUC), the best harmonic mean value (HM), the best seen accuracy (Seen), and the best unseen accuracy (Unseen), are all derived from the unseen-seen accuracy curve. We can observe that compared to other methods, our \framework consistently achieves the best trade-off between the accuracy of seen and unseen compositions, especially on the large-scale C-GQA~\cite{naeem2021learning} dataset.

\section{Ablation study with ResNet18 backbone}
\hsznew{In this paper, we use ViT as our backbone, while ResNet18 is a common choice in previous works. In~\Cref{tab:ab-backbone}, we show experimental results on ablating every component in our model with both backbones to verify the effectiveness of the proposed method. We can observe that every component is crucial for both backbones. The results indicate that our model is backbone-agnostic and performs better with ViT backbone, thanks to the capability of ViT in excavating high-level sub-space information.}
\begin{table*}[t]
\centering
\setlength{\tabcolsep}{15pt}
    \scalebox{0.75}{
    \begin{tabular}{lcccc>{\columncolor{tabcolor}}cccc>{\columncolor{tabcolor}}cccc}
        \toprule
        & & &  &  & \multicolumn{4}{c}{ViT} & \multicolumn{4}{c}{ResNet18} \\
         \cmidrule(lr){6-9} \cmidrule(lr){10-13}
         & CA & AA & OA & Reg & AUC & HM & Seen & Unseen & AUC & HM & Seen & Unseen\\
         \midrule
         (0) & \xmark & \xmark & \xmark & \xmark & 64.3 & 69.1 & 97.5 & 71.8 & 48.3 & 58.7 & 96.7 & 54.8 \\
         (1) & self & \xmark & \xmark & \xmark & 65.8 & 71.6 & 98.2 & 71.6 & 48.8 & 58.7 & 96.9 & 56.7 \\
         (2) & self & self & self & \xmark & 67.3 & \textbf{74.3} & 98.5 & 72.1 & 50.8 & 61.7 & 95.7 & 58.2 \\
         (3) & self & cross & cross & \xmark & 67.3 & 73.0 & 98.7 & 72.7 & 52.3 & 61.3 & \textbf{97.2} & 60.4 \\
         (4) & self & cross & cross & \cmark & \textbf{68.0} & 74.2 & \textbf{99.0} & \textbf{73.1} & \textbf{53.7} & \textbf{64.1} & \textbf{97.2} & \textbf{60.7}\\
         
        \bottomrule
    \end{tabular}}
    \caption{Ablate the components in \framework on open-world Clothing16K with both backbones. CA, AA, and OA denote composition, attribute, and object attention. Reg denotes the regularization term. We test self- or cross-attention for AA and OA.}
    \label{tab:ab-backbone}
\end{table*}

\section{Comparison with CNN-based models}
\begin{table*}[t]
    \centering
    \setlength{\tabcolsep}{15pt}
    \scalebox{0.75}{
    \begin{tabular}{lccccccccc}
        \toprule
         & \multicolumn{3}{c}{Composition} & \multicolumn{2}{c}{Train} & \multicolumn{2}{c}{Val} & \multicolumn{2}{c}{Test}  \\
         \cmidrule(lr){2-4} \cmidrule(lr){5-6} \cmidrule(lr){7-8} \cmidrule(lr){9-10}
         Datasets & $|\mathcal{A}|$ & $|\mathcal{O}|$ & $|\mathcal{A}|\times|\mathcal{O}|$ & $|\mathcal{C}_{s}|$ & $|\mathcal{X}|$ & $|\mathcal{C}_{s}|$ / $|\mathcal{C}_{u}|$ & $|\mathcal{X}|$ & $|\mathcal{C}_{s}|$ / $|\mathcal{C}_{u}|$ & $|\mathcal{X}|$
         \\ \midrule
        Clothing16K~\cite{zhang2022learning} & 9 & 8 & 72 & 18 & 7242 & 10 / 10 & 5515 & 9 / 8 & 3413\\
        UT-Zappos50K~\cite{yu2014fine} & 16 & 12 & 192 & 83 & 22998 & 15 / 15 & 3214 & 18 / 18 & 2914 \\
        C-GQA~\cite{naeem2021learning} & 413 & 674 & 278362 & 5592 & 26920 & 1252 / 1040 & 7280 & 888 / 923 & 5098 \\
        Vaw-CZSL~\cite{Saini_2022_CVPR} & 440 & 541 & 238040 & 11175 & 72203 & 2121 / 2322 & 9524 & 2449 / 2470 & 10856 \\
        \bottomrule
    \end{tabular}}
    \caption{Comparison of data split statistics.}
    \label{tab:data-splits-vaw}
\end{table*}

OADis~\cite{Saini_2022_CVPR} and ProtoProp~\cite{ruis2021independent} are two CZSL methods, which heavily depend on the spatial structure of convolutional features extracted from CNN models, \eg, ResNet18~\cite{he2016deep}. ProtoProp~\cite{ruis2021independent} extracts local prototypes of attribute and object features in the spatial dimension propagated through a GCN-based compositional graph. OADis~\cite{Saini_2022_CVPR} uses attribute and object affinity modules to capture the high-similarity regions in the spatial features of images with the same attribute or object. We compare our \framework with these two CNN-based models in~\cref{tab:compare-cnn}. We can observe that our model consistently outperforms other methods on all datasets. \framework increases AUC by 6.3 on Clothing16K, 1.1 on UT-Zappos59K, and 2.1 on C-GQA. In the meanwhile, \framework increases the best harmonic mean value (HM) by 4.0\% on Clothing16K, 2.3\% on UT-Zappos50K, and 4.4\% on C-GQA. The experimental results demonstrate ViT-based \framework is more efficient than the current CNN-based state-of-the-art models.
\begin{table*}[t]
    \centering
    \scalebox{0.75}{
    \begin{tabular}{l>{\columncolor{tabcolor}}cccccc>{\columncolor{tabcolor}}cccccc>{\columncolor{tabcolor}}cccccc}
        \toprule
         & \multicolumn{6}{c}{Clothing16K} & \multicolumn{6}{c}{UT-Zappos50K} & \multicolumn{6}{c}{C-GQA} \\
         \cmidrule(lr){2-7} \cmidrule(lr){8-13} \cmidrule(lr){14-19}
         Models & AUC & HM & Seen & Unseen & Attr & Obj & AUC & HM & Seen & Unseen & Attr & Obj & AUC & HM & Seen & Unseen & Attr & Obj \\
         \midrule
         ProtoProp$^\dag$~\cite{ruis2021independent} & 86.1 & 84.1 & 97.7 & 93.4 & 86.6 & 89.7 & 34.0 &  48.8 & 60.6 & \textbf{66.8} & \textbf{48.3} & 74.7 & 2.0 & 11.0 & 26.4 & 9.4 & 11.2 & 26.6\\
         OADis$^\dag$~\cite{Saini_2022_CVPR} &  85.5 & 84.7 & 96.7 & 94.1 & 84.9 & 92.5 & 30.0 & 44.4 & 59.5 & 65.5 & 46.5 & \textbf{75.5} & 3.1 & 13.6 & 30.5 & 12.7 & 10.6 & 30.7\\
         \midrule
         \framework (ours) & \textbf{92.4} & \textbf{88.7} & \textbf{98.2} & \textbf{97.7} & \textbf{90.2} & \textbf{93.6} & \textbf{35.1} & \textbf{51.1} & \textbf{63.0} & 64.3 & 46.3 & 74.0 & \textbf{5.2} & \textbf{18.0} & \textbf{35.0} & \textbf{17.7} & \textbf{16.8} & \textbf{32.3} \\ 
        \bottomrule
    \end{tabular}}
    \caption{Comparison results of \framework and two CNN-based models. We conduct experiments on Clothing16K~\cite{zhang2022learning}, UT-Zappos50K~\cite{yu2014fine}, and C-GQA~\cite{naeem2021learning} under the closed-world setting. The superscript $^\dag$ denotes the model using ResNet18~\cite{he2016deep} as the backbone.} 
    \label{tab:compare-cnn}
\end{table*}

Saini \etal~\cite{Saini_2022_CVPR} propose a new CZSL dataset, named Vaw-CZSL, a subset of Vaw~\cite{pham2021learning}, which is a multi-label attribute-object dataset. Saini \etal~\cite{Saini_2022_CVPR} sample one attribute per image, leading to a much larger dataset in comparison to previous datasets as shown in~\cref{tab:data-splits-vaw}. We compare \framework with all ViT-adapted methods in the main paper and CNN-based OADis~\cite{Saini_2022_CVPR} on Vaw-CZSL~\cite{Saini_2022_CVPR} in~\cref{tab:vaw-results}. Similar to the results on standard CZSL datasets, \framework outperforms all the other models. \framework increases AUC by 0.3 ($\sim$27.3\% relatively) and increases the best harmonic mean value (HM) by 1.2\% ($\sim$14.8\% relatively). Overall, \framework achieves stable state-of-the-art performance across various small-scale and large-scale datasets.

\begin{table}[h]
    \centering
    \setlength{\tabcolsep}{8pt}
    \scalebox{0.8}{
    \begin{tabular}{l>{\columncolor{tabcolor}}cccccc}
        \toprule
          & \multicolumn{6}{c}{Vaw-CZSL}  \\
         \cmidrule(lr){2-7} 
         Models & AUC & HM & Seen & Unseen & Attr & Obj \\
         \midrule
         SymNet~\cite{li2020symmetry} & 0.89 & 7.4 & 12.3 & 10.2 & 9.9 & 32.4 \\
         CompCos~\cite{mancini2021open} & 0.92 & 7.5 & 14.2 & 8.7 & 8.4 & 30.5\\
         GraphEmb~\cite{naeem2021learning} & 1.02 & 7.8 & 14.1 & 9.9 & 10.8 & 29.8\\
         SCEN~\cite{li2022siamese} & 0.84 & 7.1 & 14.2 & 8.1 & 7.6 & 30.0 \\ 
         IVR~\cite{zhang2022learning} & 0.91 & 7.4 & 13.0 & 9.6 & 8.9 & 31.9 \\
         OADis$^\dag$~\cite{Saini_2022_CVPR} & 0.87 & 7.1 & 13.6 & 9.4 & 9.7 & 31.4 \\
         OADis~\cite{Saini_2022_CVPR} & 1.10 & 8.1 & 15.2 & 10.1 & 9.9 & 31.6 \\
         \midrule
         \framework (ours) & \textbf{1.40} & \textbf{9.3} & \textbf{15.5} & \textbf{12.0} & \textbf{11.5} & \textbf{33.8} \\ 
        \bottomrule
    \end{tabular}}
    \caption{Experimental results on Vaw-CZSL~\cite{Saini_2022_CVPR}. We compare \framework with baseline models in the main paper and OADis~\cite{Saini_2022_CVPR}. The superscript $^\dag$ denotes the model using ResNet18~\cite{he2016deep} as the backbone. The others use ViT-B-16~\cite{dosovitskiy2020vit} as the backbone.} 
    \label{tab:vaw-results}
\end{table}

\section{Additional qualitative results}
We show additional qualitative results of \framework in this section. We follow the main paper to conduct more experiments of text-to-image retrieval, image-to-text retrieval, and visual concept retrieval, adding some results on Vaw-CZSL~\cite{Saini_2022_CVPR}.

\begin{figure*}[t]
    \centering
    \includegraphics[width=\linewidth]{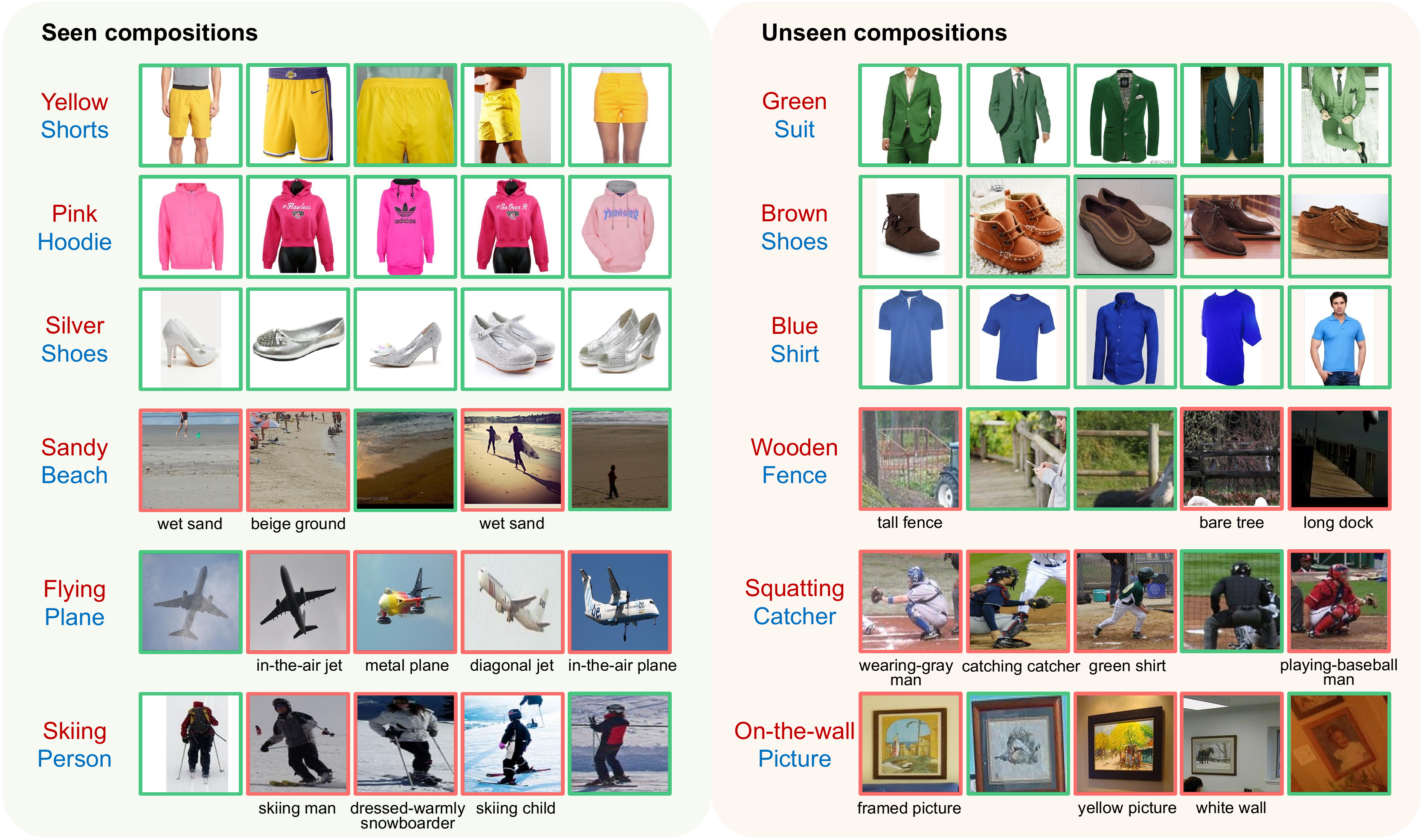}
    \caption{Text-to-image retrieval of seen (left) and unseen (right) compositions. We report the top-5 closest retrieval results on Clothing16K~\cite{zhang2022learning} (top three rows) and Vaw-CZSL~\cite{Saini_2022_CVPR} (bottom three rows). The correct image is in the green box, and the wrong image is in the red box with its ground-truth label below (black text).}
    \label{fig:supp-txt2img}
\end{figure*}

In~\cref{fig:supp-txt2img}, we retrieve the top-5 closest images for texts of attribute-object compositions. For the relatively easier Clothing16K~\cite{zhang2022learning} dataset, all the retrieved images are correct. For the more challenging large-scale Vaw-CZSL~\cite{Saini_2022_CVPR} dataset with more complicated semantics of attributes and objects, some wrong images may be retrieved but they are highly semantically-related to the given text.
Taking the ``flying plane" (row 5) as an example, the mismatched images are the ``in-the-air jet", the ``metal plane", the ``diagonal jet", and the ``in-the-air plane". These images are labelled with synonyms or from a different perspective, but they are essentially images of a ``flying plane". We can observe that \framework performs equally well for seen and unseen compositions.

\begin{figure*}[t]
    \centering
    \includegraphics[width=\linewidth]{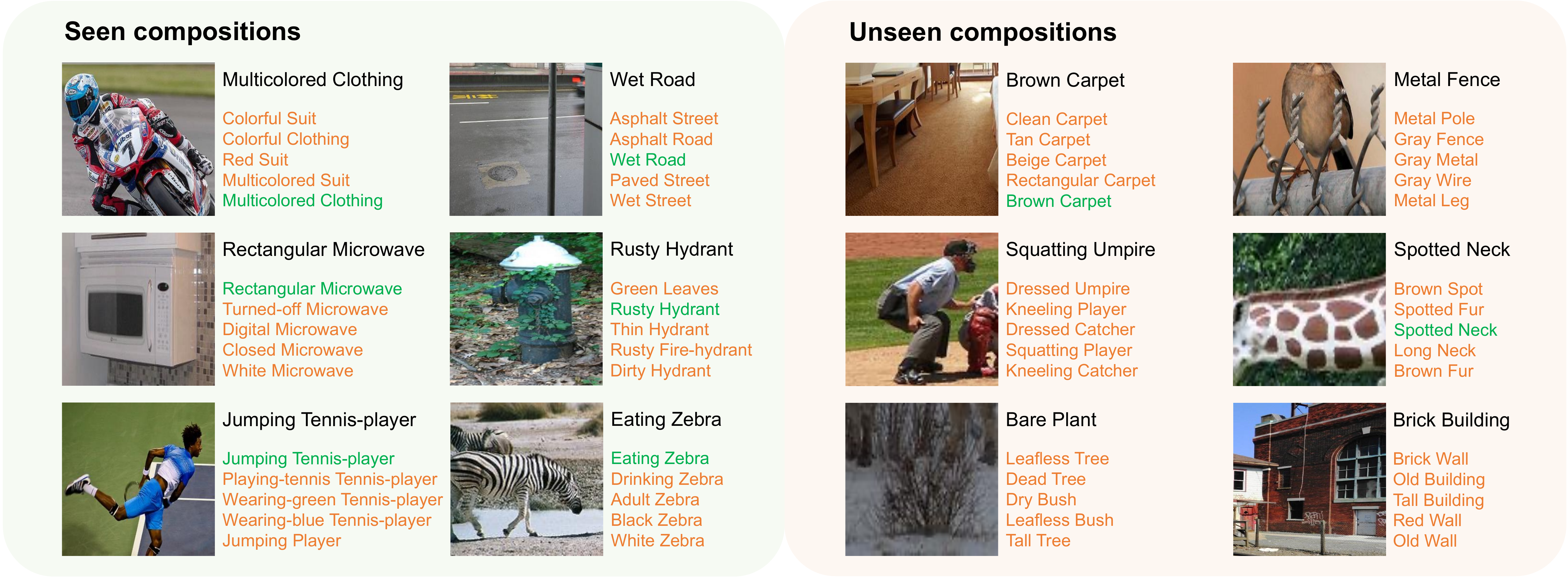}
    \caption{Image-to-text retrieval of seen (left) and unseen (right) compositions. We report the top-5 closest retrieval results on Vaw-CZSL~\cite{Saini_2022_CVPR}. The black text denotes the ground-truth label, the green text denotes the correct result, and the orange text denotes the wrong result.}
    \label{fig:supp-img2txt}
\end{figure*}

In~\cref{fig:supp-img2txt}, we retrieve the top-5 closest compositional texts for images of seen and unseen compositions. For seen compositions, it is difficult to retrieve the ground-truth label in the top-1 closest result, but all the retrieved texts are related to the image, giving the reasonable attribute-object compositions which the ground-truth label fails to incorporate. For unseen compositions, although it is quite hard to retrieve the unseen ground-truth label because of the learning bias on seen compositions, the retrieved texts are mostly reasonable to describe the given image. These results indicate \framework efficiently connects the compositional texts and the corresponding images by transferring knowledge from seen concepts to unseen compositions.

\begin{figure}[t]
    \centering
    \includegraphics[width=\linewidth]{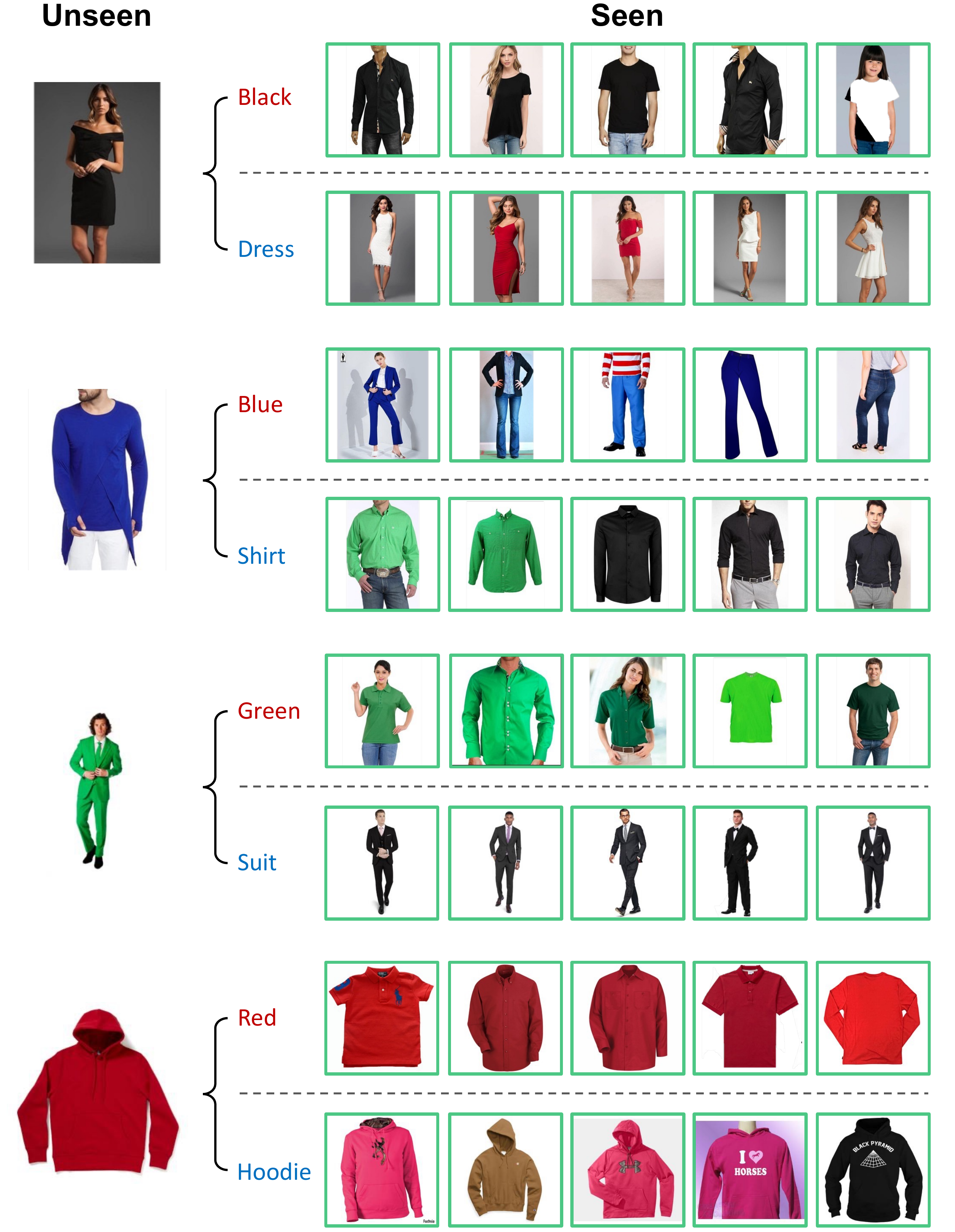}
    \caption{Retrieve \emph{seen} compositions for \emph{unseen} compositions based on the visual concept feature distance. We report the top-5 retrieval results on Clothing16K~\cite{zhang2022learning}. All the retrieved images for the corresponding concept are correct (in the green box).}
    \label{fig:supp-visual-concept}
\end{figure}

The property of \framework to disentangle concept-exclusive features enables us to conduct visual concept retrieval experiments. In~\cref{fig:supp-visual-concept}, we retrieve the attribute-related or the object-related images for the given image based on their visual concept feature distances. We report the top-5 retrieval results of four images by their attribute-exclusive and object-exclusive features. The results show that \framework effectively disentangles the attribute and object concepts from visual images and produces reliable concept-exclusive features.

\section{Pseudocode for \framework}
\framework is simple and easy to implement. For reproducibility, we show the PyTorch-style pseudocode of \framework for training in~\cref{algo:pseudocode-train} and for inference in~\cref{algo:pseudocode-infer}. 
The complete source code of ADE is available: \url{https://github.com/haoosz/ade-czsl}.
\begin{algorithm}[h]
\small
\SetAlgoLined
    \PyComment{emb\li a, emb\li o, emb\li c: embeddings of attributes, objects, and compositions} \\
    \PyComment{f: visual encoder} \\
    \PyComment{attn\li a, attn\li o, attn\li c: attribute, object, and composition attention blocks} \\
    \PyComment{proj\li a, proj\li o, proj\li c: projection with attribute, object, composition embedders} \\
    \PyComment{emd: adapted EMD at the attention level}\\
    \PyCode{} \\
    \PyComment{initialize attribute/object embeddings; compose them to composition embeddings}\\
    \PyCode{emb\li a = init(all\li attr)} \\
    \PyCode{emb\li o = init(all\li obj)} \\
    \PyCode{emb\li c = compos(emb\li a, emb\li o)} \\
    \PyComment{load 3 images and labels}\\
    \PyCode{for x, x\li a, x\li o, c in train\li loader:} \\
    \Indp   
        \PyComment{composition, attribute, object label}\\
        \PyCode{y, y\li a, y\li o = c}\\
        \PyComment{encoded tokens}\\
        \PyCode{z, z\li a, z\li o = f(x), f(x\li a), f(x\li o)}\\
        \PyComment{concept features and attention maps}\\
        \PyCode{out\li a1, amap\li aa1 = attn\li a(z, z\li a)} \\ 
        \PyCode{out\li a2, amap\li aa2 = attn\li a(z\li a, z)} \\
        \PyCode{out\li o1, amap\li oo1 = attn\li o(z, z\li o)} \\ 
        \PyCode{out\li o2, amap\li oo2 = attn\li o(z\li o, z)} \\
        \PyCode{out\li c, \li\ = attn\li c(z, z)} \\
        \PyComment{when inputs are of no interest}\\
        \PyCode{\li, amap\li ao1 = attn\li a(z, z\li o)} \\ 
        \PyCode{\li, amap\li ao2 = attn\li a(z\li o, z)} \\
        \PyCode{\li, amap\li oa1 = attn\li o(z, z\li a)} \\ 
        \PyCode{\li, amap\li oa2 = attn\li o(z\li a, z)} \\
        \PyComment{probabilities}\\
        \PyCode{p\li a1 = proj\li a(out\li a1) @ emb\li a.T}\\
        \PyCode{p\li a2 = proj\li a(out\li a2) @ emb\li a.T}\\
        \PyCode{p\li o1 = proj\li o(out\li o1) @ emb\li o.T}\\
        \PyCode{p\li o2 = proj\li o(out\li o2) @ emb\li o.T}\\
        \PyCode{p\li c = proj\li c(out\li c) @ emb\li c.T}\\
        \PyComment{cross entropy losses}\\
        \PyCode{l\li a1 = cross\li entropy(p\li a1, y\li a)}\\
        \PyCode{l\li a2 = cross\li entropy(p\li a2, y\li a)}\\
        \PyCode{l\li o1 = cross\li entropy(p\li o1, y\li o)}\\
        \PyCode{l\li o2 = cross\li entropy(p\li o2, y\li o)}\\
        \PyCode{l\li c = cross\li entropy(p\li c, y)}\\
        \PyComment{adapted EMDs}\\
        \PyCode{s\li aa = emd(amap\li aa1, amap\li aa2)}\\
        \PyCode{s\li oo = emd(amap\li oo1, amap\li oo2)}\\
        \PyCode{s\li ao = emd(amap\li ao1, amap\li ao2)}\\
        \PyCode{s\li oa = emd(amap\li oa1, amap\li oa2)}\\
        \PyComment{loss}\\
        \PyCode{l\li ce = l\li a1 + l\li a2 + l\li o1 + l\li o2 + l\li c}\\
        \PyCode{l\li reg = s\li ao + s\li oa - s\li aa - s\li oo}\\
        \PyCode{loss = l\li ce + l\li reg} \\
        \PyComment{optimization step}\\
        \PyCode{loss.backward()}\\
        \PyCode{optimizer.step()}\\
    \Indm 
\caption{PyTorch-style pseudocode for training}
\label{algo:pseudocode-train}
\end{algorithm}

\begin{algorithm}[h]
\small
\SetAlgoLined
    \PyComment{emb\li a, emb\li o, emb\li c: embeddings of attributes, objects, and compositions} \\
    \PyComment{f: visual encoder} \\
    \PyComment{attn\li a, attn\li o, attn\li c: attribute, object, and composition attention blocks} \\
    \PyComment{proj\li a, proj\li o, proj\li c: projection with attribute, object, composition embedders} \\
    \PyComment{p: a dictionary storing probabilities}\\
    \PyComment{beta: probability coefficient}\\
    \PyCode{}\\
    \PyComment{initialize attribute/object embeddings; compose them to composition embeddings}\\
    \PyCode{emb\li a = init(all\li attr)} \\
    \PyCode{emb\li o = init(all\li obj)} \\
    \PyCode{emb\li c = compos(emb\li a, emb\li o)} \\
    \PyComment{encoded tokens} \\
    \PyCode{z = f(x)}\\
    \PyComment{concept features}\\
    \PyCode{out\li a, \li\ = attn\li a(z, z)} \\ 
    \PyCode{out\li o, \li\ = attn\li o(z, z)} \\ 
    \PyCode{out\li c, \li\ = attn\li c(z, z)} \\
    \PyComment{probabilities} \\
    \PyCode{p\li a = proj\li a(out\li a) @ emb\li a.T}\\
    \PyCode{p\li o = proj\li o(out\li o) @ emb\li o.T}\\
    \PyCode{p\li c = proj\li c(out\li c) @ emb\li c.T}\\
    \PyComment{initialize an empty p} \\
    \PyCode{p = \{\}}\\
    \PyComment{enumerate all compositions} \\
    \PyCode{for c in all\li comp:} \\
    \Indp
    \PyCode{a, o = c} \PyComment{c = (a, o)}\\
    \PyComment{combine 3 probabilities} \\
    \PyCode{p[c] = p\li c[c] + beta * p\li a[a] * p\li o[o]}\\
    \Indm
    \PyCode{return p} \PyComment{return final probabilities}\\
\caption{PyTorch-style pseudocode for inference}
\label{algo:pseudocode-infer}
\end{algorithm}

\section{Broader Impacts}
Compositional zero-shot learning is a new topic of learning visual features for the objects and the corresponding attributes. Our work efficiently disentangles attribute features and object features to learn the compositionality of visual images in the real life. Our work can be used to recognize attribute-object compositions, significantly extending the traditional object recognition, which has various positive
implications, \eg, object detection, fine-grained recognition, and action recognition. Besides, our work also contributes to the explainability of deep learning models by exploring how they learn unseen things in the real world. However, there are negative impacts as well. Although it seems far from becoming true, models may be used for harmful purposes, \eg, building weapons and conducting surveillance. When the learning ability is no longer constrained by the training data and the specific training task, it is possible to train models with normal and harmless data for evil implementations because we have no idea what compositional information we can derive from the training data. In general, learning attribute and object features for compositional zero-shot learning has both positive and negative impacts, depending on how people implement this technology.

\section{Data licences}
Clothing16K\footnote{\url{https://www.kaggle.com/datasets/kaiska/apparel-dataset}}~\cite{zhang2022learning} is a split of a public dataset in Kaggle competitions under CC0 license. UT-Zappos50K\footnote{\url{https://vision.cs.utexas.edu/projects/finegrained/utzap50k/}} is collected by Yu \etal~\cite{yu2014fine}, allowing non-commercial research use. C-GQA~\cite{naeem2021learning} is a split built on top of Stanford GQA dataset\footnote{\url{https://cs.stanford.edu/people/dorarad/gqa/index.html}}~\cite{hudson2019gqa}, which is free for non-commercial research use. 
Vaw-CZSL~\cite{Saini_2022_CVPR} is a subset of Vaw\footnote{\url{https://vawdataset.com/}}~\cite{pham2021learning} under MIT license. 

%% file: _main.bbl
\begin{thebibliography}{10}\itemsep=-1pt

\bibitem{alayrac2017joint}
Jean-Baptiste Alayrac, Ivan Laptev, Josef Sivic, and Simon Lacoste-Julien.
\newblock Joint discovery of object states and manipulation actions.
\newblock In {\em ICCV}, 2017.

\bibitem{atzmon2020causal}
Yuval Atzmon, Felix Kreuk, Uri Shalit, and Gal Chechik.
\newblock A causal view of compositional zero-shot recognition.
\newblock In {\em NeurIPS}, 2020.

\bibitem{caron2021emerging}
Mathilde Caron, Hugo Touvron, Ishan Misra, Herv\'e J\'egou, Julien Mairal,
  Piotr Bojanowski, and Armand Joulin.
\newblock Emerging properties in self-supervised vision transformers.
\newblock In {\em ICCV}, 2021.

\bibitem{chao2016empirical}
Wei-Lun Chao, Soravit Changpinyo, Boqing Gong, and Fei Sha.
\newblock An empirical study and analysis of generalized zero-shot learning for
  object recognition in the wild.
\newblock In {\em ECCV}, 2016.

\bibitem{chen2014inferring}
Chao-Yeh Chen and Kristen Grauman.
\newblock Inferring analogous attributes.
\newblock In {\em CVPR}, 2014.

\bibitem{deng2009imagenet}
Jia Deng, Wei Dong, Richard Socher, Li-Jia Li, Kai Li, and Li Fei-Fei.
\newblock Imagenet: A large-scale hierarchical image database.
\newblock In {\em CVPR}, 2009.

\bibitem{dosovitskiy2020vit}
Alexey Dosovitskiy, Lucas Beyer, Alexander Kolesnikov, Dirk Weissenborn,
  Xiaohua Zhai, Thomas Unterthiner, Mostafa Dehghani, Matthias Minderer, Georg
  Heigold, Sylvain Gelly, Jakob Uszkoreit, and Neil Houlsby.
\newblock An image is worth 16x16 words: Transformers for image recognition at
  scale.
\newblock In {\em ICLR}, 2021.

\bibitem{farhadi2010attribute}
Ali Farhadi, Ian Endres, and Derek Hoiem.
\newblock Attribute-centric recognition for cross-category generalization.
\newblock In {\em CVPR}, 2010.

\bibitem{farhadi2009describing}
Ali Farhadi, Ian Endres, Derek Hoiem, and David Forsyth.
\newblock Describing objects by their attributes.
\newblock In {\em CVPR}, 2009.

\bibitem{fathi2013modeling}
Alireza Fathi and James~M Rehg.
\newblock Modeling actions through state changes.
\newblock In {\em CVPR}, 2013.

\bibitem{ferrari2007learning}
Vittorio Ferrari and Andrew Zisserman.
\newblock Learning visual attributes.
\newblock {\em NeurIPS}, 2007.

\bibitem{goodfellow2014generative}
Ian~J Goodfellow, Jean Pouget-Abadie, Mehdi Mirza, Bing Xu, David Warde-Farley,
  Sherjil Ozair, Aaron~C Courville, and Yoshua Bengio.
\newblock Generative adversarial nets.
\newblock In {\em NeurIPS}, 2014.

\bibitem{he2016deep}
Kaiming He, Xiangyu Zhang, Shaoqing Ren, and Jian Sun.
\newblock Deep residual learning for image recognition.
\newblock In {\em CVPR}, 2016.

\bibitem{emd_paper}
Frank~L. Hitchcock.
\newblock The distribution of a product from several sources to numerous
  localities.
\newblock {\em Journal of Mathematics and Physics}, 1941.

\bibitem{hudson2019gqa}
Drew~A Hudson and Christopher~D Manning.
\newblock Gqa: A new dataset for real-world visual reasoning and compositional
  question answering.
\newblock In {\em CVPR}, 2019.

\bibitem{hwang2011sharing}
Sung~Ju Hwang, Fei Sha, and Kristen Grauman.
\newblock Sharing features between objects and their attributes.
\newblock In {\em CVPR}, 2011.

\bibitem{isola2015discovering}
Phillip Isola, Joseph~J Lim, and Edward~H Adelson.
\newblock Discovering states and transformations in image collections.
\newblock In {\em CVPR}, 2015.

\bibitem{karthik2022kg}
Shyamgopal Karthik, Massimiliano Mancini, and Zeynep Akata.
\newblock Kg-sp: Knowledge guided simple primitives for open world
  compositional zero-shot learning.
\newblock In {\em CVPR}, 2022.

\bibitem{kingma2015adam}
Diederik~P Kingma and Jimmy Ba.
\newblock Adam: A method for stochastic optimization.
\newblock In {\em ICLR}, 2015.

\bibitem{kulkarni2013babytalk}
Girish Kulkarni, Visruth Premraj, Vicente Ordonez, Sagnik Dhar, Siming Li,
  Yejin Choi, Alexander~C Berg, and Tamara~L Berg.
\newblock Babytalk: Understanding and generating simple image descriptions.
\newblock {\em IEEE TPAMI}, 2013.

\bibitem{lampert2009learning}
Christoph~H Lampert, Hannes Nickisch, and Stefan Harmeling.
\newblock Learning to detect unseen object classes by between-class attribute
  transfer.
\newblock In {\em CVPR}, 2009.

\bibitem{li2022siamese}
Xiangyu Li, Xu Yang, Kun Wei, Cheng Deng, and Muli Yang.
\newblock Siamese contrastive embedding network for compositional zero-shot
  learning.
\newblock In {\em CVPR}, 2022.

\bibitem{li2020symmetry}
Yong-Lu Li, Yue Xu, Xiaohan Mao, and Cewu Lu.
\newblock Symmetry and group in attribute-object compositions.
\newblock In {\em CVPR}, 2020.

\bibitem{mahajan2011joint}
Dhruv Mahajan, Sundararajan Sellamanickam, and Vinod Nair.
\newblock A joint learning framework for attribute models and object
  descriptions.
\newblock In {\em ICCV}, 2011.

\bibitem{mancini2021open}
M Mancini, MF Naeem, Y Xian, and Zeynep Akata.
\newblock Open world compositional zero-shot learning.
\newblock In {\em CVPR}, 2021.

\bibitem{mancini2022learning}
Massimiliano Mancini, Muhammad~Ferjad Naeem, Yongqin Xian, and Zeynep Akata.
\newblock Learning graph embeddings for open world compositional zero-shot
  learning.
\newblock {\em IEEE TPAMI}, 2022.

\bibitem{mccandless2013object}
Tomas McCandless and Kristen Grauman.
\newblock Object-centric spatio-temporal pyramids for egocentric activity
  recognition.
\newblock In {\em BMVC}, 2013.

\bibitem{mikolov2013distributed}
Tomas Mikolov, Ilya Sutskever, Kai Chen, Greg~S Corrado, and Jeff Dean.
\newblock Distributed representations of words and phrases and their
  compositionality.
\newblock In {\em NeurIPS}, 2013.

\bibitem{misra2017red}
Ishan Misra, Abhinav Gupta, and Martial Hebert.
\newblock From red wine to red tomato: Composition with context.
\newblock In {\em CVPR}, 2017.

\bibitem{naeem2021learning}
MF Naeem, Y Xian, F Tombari, and Zeynep Akata.
\newblock Learning graph embeddings for compositional zero-shot learning.
\newblock In {\em CVPR}, 2021.

\bibitem{nagarajan2018attributes}
Tushar Nagarajan and Kristen Grauman.
\newblock Attributes as operators: factorizing unseen attribute-object
  compositions.
\newblock In {\em ECCV}, 2018.

\bibitem{nan2019recognizing}
Zhixiong Nan, Yang Liu, Nanning Zheng, and Song-Chun Zhu.
\newblock Recognizing unseen attribute-object pair with generative model.
\newblock In {\em AAAI}, 2019.

\bibitem{nayak2022learning}
Nihal~V Nayak, Peilin Yu, and Stephen~H Bach.
\newblock Learning to compose soft prompts for compositional zero-shot
  learning.
\newblock {\em arXiv preprint arXiv:2204.03574}, 2022.

\bibitem{ordonez2011im2text}
Vicente Ordonez, Girish Kulkarni, and Tamara Berg.
\newblock Im2text: Describing images using 1 million captioned photographs.
\newblock In {\em NeurIPS}, 2011.

\bibitem{palatucci2009zero}
Mark Palatucci, Dean Pomerleau, Geoffrey~E Hinton, and Tom~M Mitchell.
\newblock Zero-shot learning with semantic output codes.
\newblock In {\em NeurIPS}, 2009.

\bibitem{patterson2016coco}
Genevieve Patterson and James Hays.
\newblock Coco attributes: Attributes for people, animals, and objects.
\newblock In {\em ECCV}, 2016.

\bibitem{pham2021learning}
Khoi Pham, Kushal Kafle, Zhe Lin, Zhihong Ding, Scott Cohen, Quan Tran, and
  Abhinav Shrivastava.
\newblock Learning to predict visual attributes in the wild.
\newblock In {\em CVPR}, 2021.

\bibitem{purushwalkam2019task}
Senthil Purushwalkam, Maximilian Nickel, Abhinav Gupta, and Marc'Aurelio
  Ranzato.
\newblock Task-driven modular networks for zero-shot compositional learning.
\newblock In {\em ICCV}, 2019.

\bibitem{radford2021learning}
Alec Radford, Jong~Wook Kim, Chris Hallacy, Aditya Ramesh, Gabriel Goh,
  Sandhini Agarwal, Girish Sastry, Amanda Askell, Pamela Mishkin, Jack Clark,
  et~al.
\newblock Learning transferable visual models from natural language
  supervision.
\newblock In {\em ICML}, 2021.

\bibitem{raghu2021vision}
Maithra Raghu, Thomas Unterthiner, Simon Kornblith, Chiyuan Zhang, and Alexey
  Dosovitskiy.
\newblock Do vision transformers see like convolutional neural networks?
\newblock In {\em NeurIPS}, 2021.

\bibitem{romera2015embarrassingly}
Bernardino Romera-Paredes and Philip Torr.
\newblock An embarrassingly simple approach to zero-shot learning.
\newblock In {\em ICML}, 2015.

\bibitem{ruis2021independent}
Frank Ruis, Gertjan Burghouts, and Doina Bucur.
\newblock Independent prototype propagation for zero-shot compositionality.
\newblock In {\em NeurIPS}, 2021.

\bibitem{Saini_2022_CVPR}
Nirat Saini, Khoi Pham, and Abhinav Shrivastava.
\newblock Disentangling visual embeddings for attributes and objects.
\newblock In {\em CVPR}, 2022.

\bibitem{shrivastava2012constrained}
Abhinav Shrivastava, Saurabh Singh, and Abhinav Gupta.
\newblock Constrained semi-supervised learning using attributes and comparative
  attributes.
\newblock In {\em ECCV}, 2012.

\bibitem{socher2013zero}
Richard Socher, Milind Ganjoo, Christopher~D Manning, and Andrew Ng.
\newblock Zero-shot learning through cross-modal transfer.
\newblock In {\em NeurIPS}, 2013.

\bibitem{vaswani2017attention}
Ashish Vaswani, Noam Shazeer, Niki Parmar, Jakob Uszkoreit, Llion Jones,
  Aidan~N Gomez, {\L}ukasz Kaiser, and Illia Polosukhin.
\newblock Attention is all you need.
\newblock In {\em NeurIPS}, 2017.

\bibitem{wang2018non}
Xiaolong Wang, Ross Girshick, Abhinav Gupta, and Kaiming He.
\newblock Non-local neural networks.
\newblock In {\em CVPR}, 2018.

\bibitem{wei2019adversarial}
Kun Wei, Muli Yang, Hao Wang, Cheng Deng, and Xianglong Liu.
\newblock Adversarial fine-grained composition learning for unseen
  attribute-object recognition.
\newblock In {\em ICCV}, 2019.

\bibitem{xian2018feature}
Yongqin Xian, Tobias Lorenz, Bernt Schiele, and Zeynep Akata.
\newblock Feature generating networks for zero-shot learning.
\newblock In {\em CVPR}, 2018.

\bibitem{xian2017zero}
Yongqin Xian, Bernt Schiele, and Zeynep Akata.
\newblock Zero-shot learning-the good, the bad and the ugly.
\newblock In {\em CVPR}, 2017.

\bibitem{yu2014fine}
Aron Yu and Kristen Grauman.
\newblock Fine-grained visual comparisons with local learning.
\newblock In {\em CVPR}, 2014.

\bibitem{zhang2022learning}
Tian Zhang, Kongming Liang, Ruoyi Du, Xian Sun, Zhanyu Ma, and Jun Guo.
\newblock Learning invariant visual representations for compositional zero-shot
  learning.
\newblock In {\em ECCV}, 2022.

\end{thebibliography}
